%% file: iclr2024_conference.tex
\documentclass{article} % For LaTeX2e
\usepackage{iclr2024_conference,times}

\input{math_commands.tex}

\usepackage{hyperref}
\usepackage{url}

\usepackage{graphicx}
\usepackage[inline]{enumitem}
\usepackage{amsmath}
\usepackage{booktabs}
\usepackage{multirow}
\usepackage{color}
\usepackage{makecell}
\usepackage{xcolor}

\usepackage{subfig}
\usepackage{float}
\usepackage{hyperref}
\usepackage{pifont}

\def\eg{\emph{e.g.}} 
\def\ie{\emph{i.e.}}

\usepackage{wrapfig}

\title{AnomalyCLIP: Object-agnostic Prompt Learning for Zero-shot Anomaly Detection}

\author{    % Authors
    Qihang	Zhou\textsuperscript{\rm 1}\thanks{Equal contribution. $\dagger$ Corresponding authors.} ,
    Guansong Pang\textsuperscript{\rm 2$\ast$},
    Yu	Tian\textsuperscript{\rm 3},
    Shibo	He\textsuperscript{\rm 1$\dagger$},
    Jiming	Chen\textsuperscript{\rm 1$\dagger$} \\
\textsuperscript{\rm 1}Zhejiang University \quad \textsuperscript{\rm 2} Singapore Management University \quad \textsuperscript{\rm 3}Harvard University \\
\textsuperscript{\rm 1} \texttt{\{zqhang, s18he, cjm\}@zju.edu.cn} \quad \textsuperscript{\rm 2}\texttt{gspang@smu.edu.sg} \quad \\ \textsuperscript{\rm 3}\texttt{ytian11@meei.harvard.edu}
}

\iclrfinaltrue
\begin{document}

\maketitle

\begin{abstract}
Zero-shot anomaly detection (ZSAD) requires detection models trained using auxiliary data to detect anomalies without any training sample in a target dataset. It is a crucial task when training data is not accessible due to various concerns, \eg, data privacy, yet it is challenging since the models need to generalize to anomalies across different domains where the appearance of foreground objects, abnormal regions, and background features, such as defects/tumors on different products/organs, can vary significantly.
Recently large pre-trained vision-language models (VLMs), such as CLIP,
have demonstrated strong zero-shot recognition ability in various vision tasks, including anomaly detection. However, their ZSAD performance is weak since the VLMs focus more on modeling the class semantics of the foreground objects rather than the abnormality/normality in the images.
In this paper we introduce a novel approach, namely AnomalyCLIP, to adapt CLIP for accurate ZSAD across different domains. The key insight of AnomalyCLIP is to learn object-agnostic text prompts that capture generic normality and abnormality in an image regardless of its foreground objects. This allows our model to focus on the abnormal image regions rather than the object semantics, enabling generalized normality and abnormality recognition on diverse types of objects. Large-scale experiments on 17 real-world anomaly detection datasets show that AnomalyCLIP achieves superior zero-shot performance of detecting and segmenting anomalies in datasets of highly diverse class semantics from various defect inspection and medical imaging domains.  Code will be made available at \url{https://github.com/zqhang/AnomalyCLIP}.
\end{abstract}

\section{Introduction}
\begin{figure*}
  \centering
  \resizebox{1\textwidth}{!}{
    \subfloat[]{\includegraphics[width=0.16\textwidth]{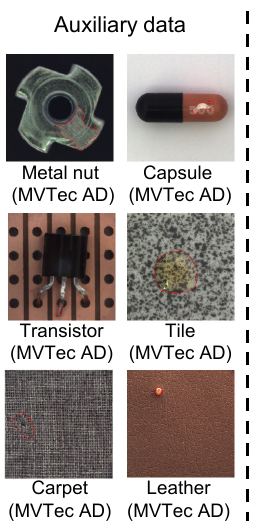}%
    \label{fig: class_comparison_revise 1}}
    \hfil
    \subfloat[]{\includegraphics[width=0.16\textwidth]{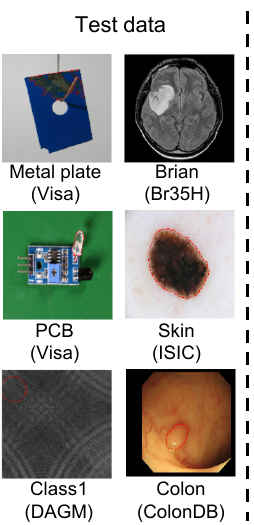}%
    \label{fig: class_comparison_revise 2}}
    \hfil
    \subfloat[]{\includegraphics[width=0.16\textwidth]{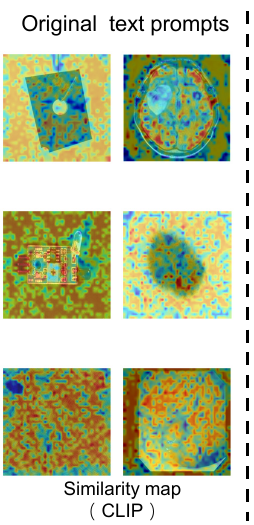}%
    \label{fig: class_comparison_revise 3}}
    \hfil
    \subfloat[]{\includegraphics[width=0.16\textwidth]{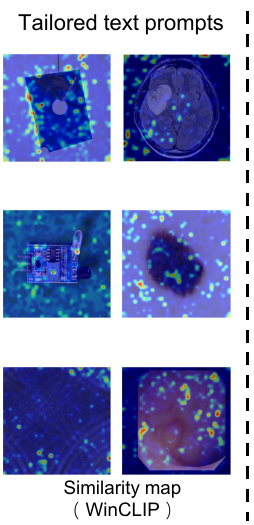}%
    \label{fig: class_comparison_revise 4}}
    \hfil
    \subfloat[]{\includegraphics[width=0.16\textwidth]{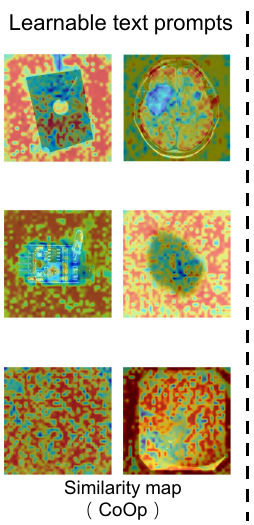}%
    \label{fig: class_comparison_revise 5}}
    \hfil
    \subfloat[]{\includegraphics[width=0.16\textwidth]{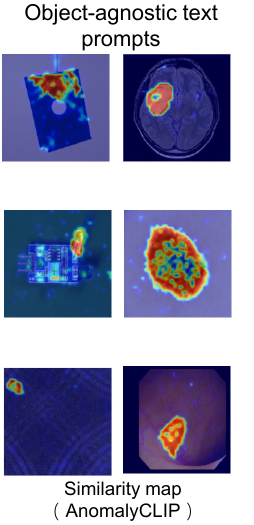}%
    \label{fig: class_comparison_revise 6}}}
    \hfil
    % \vspace{-0.8em}
 \caption{Comparison of ZSAD results on (b) test data using (c) original text prompts in CLIP \citep{radford2021learning}, (d) tailored text prompts for AD in WinCLIP \citep{jeong2023winclip}, (e) learnable text prompts for general vision tasks in CoOp \citep{zhou2022conditional}, and (f) object-agnostic text prompts in our AnomalyCLIP. (a) presents a set of auxiliary data we can use to learn the text prompts. The results are obtained by measuring the similarity between text prompt embeddings and image embeddings. The ground-truth anomaly regions are circled in red in (a) and (b). (c), (d), and (e) suffer from poor generalization across different domains, while our AnomalyCLIP in (f) can well generalize to anomalies in diverse types of objects from different domains.}

\vspace{-1em}
 \label{fig:auroc_curves}
\end{figure*}

Anomaly detection (AD) has been widely applied in various applications, such as industrial defect inspection~\citep{bergmann2019mvtec, xie2023pushing, roth2022towards, huang2022registration, mou2022rgi, chen2022deep, bergmann2020uninformed,pang2021explainable, reiss2023mean, you2022unified, liznerski2020explainable,ding2022catching, 9940966, cao2023anomaly} and medical image analysis~\citep{pang2021explainable,qin2022medical,liu2023clip,ding2022catching,tian2021constrained,tian2023self,fernando2021deep}. 
Existing AD approaches typically assume that training examples in a target application domain are available for learning the detection models \citep{pang2021deep,ruff2021unifying}. However, this assumption may not hold in various scenarios, such as i) when accessing training data violates data privacy policies (\eg, to protect the sensitive information of patients), or ii) when the target domain does not have relevant training data (\eg, 
inspecting defects in a manufacturing line of new products). Zero-shot anomaly detection (ZSAD) is an emerging task for AD in such scenarios, to which the aforementioned AD approaches are not viable, as it requires detection models to detect anomalies without any training sample in a target dataset. Since anomalies from different application scenarios typically have substantial variations in their visual appearance, foreground objects, and background features, \eg, defects on the surface of one product vs. that on the other products, lesions/tumors on different organs, or industrial defects vs. tumors/lesions in medical images, detection models with strong generalization ability w.r.t. such variations are needed for accurate ZSAD. Recently large pre-trained vision-language models (VLMs)~\citep{radford2021learning, kirillov2023segment} have demonstrated strong zero-shot recognition ability in various vision tasks, including anomaly detection \citep{jeong2023winclip}. Particularly, being pre-trained using millions/billions of image-text pairs, CLIP~\citep{radford2021learning} 
has been applied to empower various downstream tasks~\citep{zhou2022learning, rao2022denseclip, khattak2023maple, sain2023clip} with its strong generalization capability. WinCLIP \citep{jeong2023winclip} is a seminal work in the ZSAD line, which designs a large number of artificial text prompts to exploit the CLIP's generalizability for ZSAD. However, the VLMs such as CLIP are primarily trained to align with the class semantics of foreground objects rather than the abnormality/normality in the images, and as a result, their generalization in understanding the visual abnormality/normality is restricted, leading to weak ZSAD performance. Further, the current prompting approaches, using either manually defined text prompts \citep{jeong2023winclip} or learnable prompts~\citep{sun2022dualcoop, zhou2022conditional}, often result in prompt embeddings that opt for global features for effective object semantic alignment
\citep{zhong2022regionclip,wu2023aligning}, failing to capture the abnormality that often manifests in fine-grained, local features, as shown in Fig.~\ref{fig: class_comparison_revise 4} and Fig.~\ref{fig: class_comparison_revise 5}.
In this paper we introduce a novel approach, namely AnomalyCLIP, to adapt CLIP for accurate ZSAD across different domains. AnomalyCLIP aims to learn object-agnostic text prompts that capture generic normality and abnormality in an image regardless of its foreground objects. It first devises a simple yet universally-effective learnable prompt template for the two general classes -- normality and abnormality -- and then utilizes both image-level and pixel-level loss functions to learn the generic normality and abnormality globally and locally in our prompt embeddings using auxiliary data. This allows our model to focus on the abnormal image regions rather the object semantics, enabling remarkable zero-shot capability of recognizing the abnormality that has similar abnormal patterns to those in auxiliary data. 
% The common
As shown in Fig.~\ref{fig: class_comparison_revise 1} and Fig.~\ref{fig: class_comparison_revise 2}, the foreground object semantics can be completely different in the fine-tuning auxiliary data and target data, but the anomaly patterns remain similar, \eg, scratches on metal nuts and plates, the misplacement of transistors and PCB, tumors/lesions on various organ surfaces, etc. Text prompt embeddings in CLIP fail to generalize across different domains, as illustrated in Fig.~\ref{fig: class_comparison_revise 3}, but object-agnostic prompt embeddings learned by AnomalyCLIP can effectively generalize to recognize the abnormality across different domain images in Fig.~\ref{fig: class_comparison_revise 6}.
    
In summary, this paper makes the following main contributions.
\begin{itemize}
    \item We reveal for the first time that learning object-agnostic text prompts of normality and abnormality is a simple yet effective approach for accurate ZSAD. Compared to current text prompting approaches that are primarily designed for object semantic alignment \citep{jeong2023winclip,zhou2022learning}, our text prompt embeddings model semantics of generic abnormality and normality, allowing object-agnostic, generalized ZSAD performance.
    \item We then introduce a novel ZSAD approach, called AnomalyCLIP, in which we utilize an object-agnostic prompt template and a glocal abnormality loss function (i.e., a combination of global and local loss functions) to learn the generic abnormality and normality prompts using auxiliary data. In doing so, AnomalyCLIP largely simplifies the prompt design and can effectively apply to different domains without requiring any change on its learned two prompts, contrasting to existing methods like WinCLIP whose effectiveness relies heavily on extensive engineering on hundreds of manually defined prompts.
    \item Comprehensive experiments on 17 datasets from various industrial and medical domains demonstrate that AnomalyCLIP achieves superior ZSAD performance of detecting and segmenting anomalies in datasets of highly diverse class semantics from defect inspection and medical imaging domains.

\end{itemize}

% \vspace{-1.5em}
\section{Preliminary}
% \vspace{-0.5em}
CLIP consists of a text encoder and visual encoder denoted as $T(\cdot)$ and $F(\cdot)$, respectively. Both encoders are mainstream multi-layer networks such as ViT~\citep{dosovitskiy2020image, vaswani2017attention}. Using text prompts is a typical way to achieve the embeddings of different classes for zero-shot recognition. Particularly, a text prompt template $\sG$ with the class name $c$ can be passed through $T(\cdot)$ to obtain its corresponding textual embedding $g_c \in \mathbb{R}^D$.
The text prompt template commonly used in CLIP looks like \verb'A photo of a [cls]', where \verb'[cls]' represents the target class name. 
% As for visual encoder,
Then $F(\cdot)$ encodes an image $x_i$ to derive visual representations, where the class token $f_i \in \R ^ D$ is treated as its visual embedding (global visual embedding), and patch tokens $f_i^m \in \R ^ {H\times W\times D}$ are referred to as local visual embeddings. 
% The intermediate visual feature map of n-th is denoted as $\vf_{i, n}^m$. 
CLIP performs zero-shot recognition by measuring the similarity between textual and visual embeddings. In specific, given a target class set $\mathcal{C}$ and an image $x_i$, CLIP predicts the probability of $x_i$ belonging to $c$ as follows: 
\vspace{-1pt}
\begin{equation}
     p(y = c | x_i) = P(g_c, f_i) = {\frac{exp(<g_c, f_i>/\tau)}{\sum_{c\in \mathcal{C}}exp(<g_c, f_i>)/\tau)}},
\label{equ: softmax}
\end{equation}
\vspace{-2pt}
where $\tau$ is a temperature hyperparameter, and the operator $<\cdot, \cdot>$ represents the computation of cosine similarity. Unlike many vision tasks that involve many objects and use the name of the objects as the class name \verb'[cls]',
% Consider 
we posit that performing ZSAD tasks using CLIP should be object-agnostic, so
% , a natural solution is 
we propose to design two classes of text prompts (\ie, normality and abnormality) and compute the possibility of these two classes according to Eq.~\ref{equ: softmax}.
We denote the probability of being abnormal $P(g_a, f_i)$ as the anomaly score. The computation is extended from global visual embeddings to local visual embeddings to derive the corresponding segmentation maps $S_n\in \R ^ {H\times W}$ and $S_a  \in \R ^ {H\times W}$, where each entry $(j,k)$ are computed as $P(g_n, f^{m(j, k)}_i)$ and $P(g_a, f^{m(j, k)}_i)$.
\begin{figure*}[t]
    \centering
    \includegraphics[width=1\textwidth]{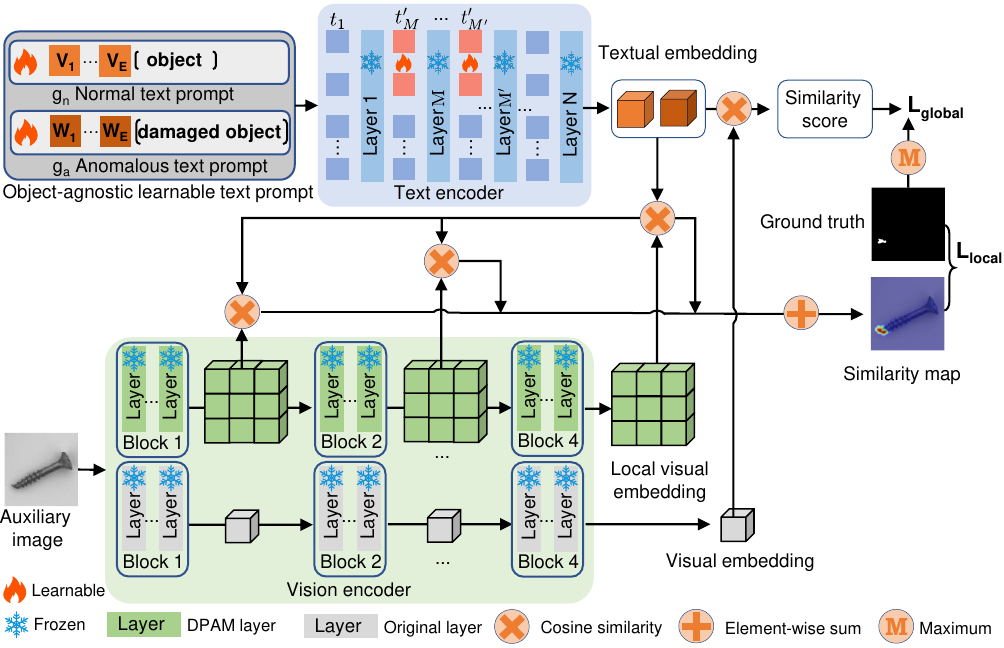}
    \caption{Overview of AnomalyCLIP. To adapt CLIP to ZSAD, AnomalyCLIP introduces object-agnostic text prompt templates to capture generic normality and abnormality regardless of the object semantics. Then, we introduce glocal context optimization to incorporate global and fine-grained anomaly semantics into object-agnostic text prompt learning. Finally, textual prompt tuning and DPAM are used to enable the prompt learning in the textual and local visual spaces of CLIP.}
    % \vspace{-1.5em}
    \label{fig1:overview}
\end{figure*}

% \vspace{-0.5em}
\section{AnomalyCLIP: object-agnostic prompt learning}
\subsection{Approach overview}
In this paper, we propose AnomalyCLIP to adapt CLIP to ZSAD via object-agnostic prompt learning. As shown in Fig.~\ref{fig1:overview}, AnomalyCLIP first introduces object-agnostic text prompt templates, where we design two generic object-agnostic text prompt templates of $g_n$ and $g_a$ to learn generalized embedding for the normality and abnormality classes, respectively (see Sec.~\ref{The design of learnable text prompts}). To learn such generic text prompt templates, we introduce global and local context optimization to incorporate global and fine-grained anomaly semantics into object-agnostic textual embedding learning. In addition, textual prompt tuning and DPAM are used to support the learning in the textual and local visual spaces of CLIP. Finally, we could integrate the multiple intermediate layers to provide more local visual details~\footnote{Although we use only the top feature map of visual encoder in the code implementation, AnomalyCLIP is capable of incorporating multiple intermediate feature maps.}. During training, all modules are jointly optimized by the combination of global and local context optimization. During inference, we quantify the misalignment of textual and global/local visual embeddings to obtain the anomaly score and anomaly score map, respectively (see Sec.~\ref{global and local context optimization}).

\subsection{Object-agnostic text prompt design}
\label{The design of learnable text prompts}

Commonly used text prompt templates in CLIP, like \verb'A photo of a [cls]', primarily focus on object semantics. Consequently, they fail to generate textual embeddings that capture anomaly and normal semantics to query corresponding visual embeddings. To support the learning of anomaly-discriminative textual embeddings, we aim to incorporate prior anomaly semantics into text prompt templates. A trivial solution is to design the templates with specific anomaly types, such as \verb' A photo of a [cls] with scratches'
% in WinCLIP \citep{jeong2023winclip}. 
However, the pattern of anomaly is typically unknown and diverse, so it is practically difficult to list all possible anomaly types. Therefore, it is important to define text prompt templates with generic anomaly semantics. For this purpose, we can adopt the text \verb'damaged [cls]' to cover comprehensive anomaly semantics, facilitating the detection of diverse defects such as scratches and holes. Nevertheless, utilizing such text prompt templates poses challenges in generating generic anomaly-discriminating textual embeddings. This is because CLIP's original pre-training focuses on aligning with object semantics instead of the abnormality and normality within images. To address this limitation, we can introduce learnable text prompt templates and tune the prompts using auxiliary AD-relevant data. During the fine-tuning process, these learnable templates can incorporate both broad and detailed anomaly semantics, resulting in textual embeddings that are more discriminative between normality and abnormality. This helps avoid the need for manually defined text prompt templates that require extensive engineering~\citep{jeong2023winclip}. 
These text prompts are referred to as \textbf{object-aware text prompt templates} and defined as follows:
\begin{align}
    g_n &= [V_1][V_2]\dots[V_E][cls] \nonumber\\
    g_a &= [W_1][W_2]\dots[W_E] [damaged][cls],\nonumber
\end{align}
where $[V]_i$ and $[W]_i$ ($i \in {1, \dots, E}$) are learnable word embeddings in normality and abnormality text prompt templates, respectively.

ZSAD tasks require models to detect anomalies in previously unseen target datasets. These datasets often exhibit significant variations in object semantics among different objects, like various defects on one product vs. another, or discrepancies between industrial defects and medical imaging tumors. However, despite these substantial differences in object semantics, the underlying anomaly patterns could be similar. For instance, anomalies like scratches on metal nuts and plates, or the misplacement of transistors and PCB, as well as tumors on the surface of various organs, can share similar anomaly patterns. We hypothesize that the key of accurate ZSAD is to identify these generic anomaly patterns regardless of the varying semantics of different objects. Therefore, the inclusion of object semantics in object-aware text prompt templates is often unnecessary for ZSAD. It can even hinder the detection of anomalies in classes that have not been seen during the learning process. More importantly, excluding the object semantics from text prompt templates allows learnable text prompt templates to focus on capturing the characteristics of anomalies themselves, rather than the objects. Motivated by this, we introduce object-agnostic prompt learning, with the aim to capture generic normality and abnormality within images regardless of the object semantics. Different from object-aware text prompt templates, as shown below, the \textbf{object-agnostic text prompt templates} replace the class name in $g_n$ and $g_a$ with \verb'object', blocking out the class semantics of objects: 
\begin{align}
    g_n&= [V_1][V_2]\dots[V_E][object] \nonumber\\
    g_a &= [W_1][W_2]\dots[W_E][damaged][object].\nonumber
\end{align}
This design empowers the object-agnostic text prompt template to learn the shared patterns of different anomalies. As a result, the generated textual embeddings are more generic and capable of identifying anomalies across diverse objects and different domains. Further, this prompt design is versatile and can be applied to different target domains without any modification, \eg, requiring no knowledge about the object name or anomaly types in a target dataset.

\subsection{Learning generic abnormality and normality prompts}
% \subsection{Global and local context optimization}
\label{global and local context optimization}
\paragraph{Glocal context optimization} 
To effectively learn the object-agnostic text prompts, we devise a joint optimization approach that enables the normality and abnormality prompt learning from both global and local perspectives, namely global and local context optimization. 
% To align object-agnostic textual and global visual embeddings, we introduce 
The global context optimization
% . Here, textual embeddings that 
% are
aims to enforce that our object-agnostic textual embeddings are matched with the global visual embeddings of images of diverse objects. This helps effectively capture the normal/abnormal semantics from a global feature perspective. The local context optimization is introduced to enable object-agnostic text prompts to concentrate on fine-grained, local abnormal regions from $M$ intermediate layers of the visual encoder, in addition to the global normal/abnormal features. Formally, let $\mathcal{M}$ be the set of intermediate layers used (\ie, $M=|\mathcal{M}|$), our text prompts are learned by minimizing the following glocal loss function:
\begin{align}\label{eq:total_loss}
    L_{total} = L_{global} + \lambda \textstyle \sum_{M_l \in \mathcal{M}}L_{local}^{M_l},
\end{align}
where $\lambda$ is a hyperparameter to balance the global and local losses. $L_{global}$ is a cross-entropy loss that matches the cosine similarity between the object-agnostic textual embeddings and visual embeddings of normal/abnormal images from auxiliary data. Let $S \in \R^{H_{image} \times W_{image}}$ be the ground-truth segmentation mask, with $S_{jk} = 1$ if the pixel is as an anomaly and $S_{jk} = 0$ otherwise, then we have
\vspace{-1em}
\begin{gather}
\color{black}
  S^{(j,k)}_{n, M_l} = P(g_n, f^{m(j, k)}_{i, M_l}), \enspace
    \color{black}
    S^{(j,k)}_{a, M_l} = \textstyle P(g_a, f^{m(j, k)}_{i, M_l}), \enspace \text{where} \enspace j\in[1, H], k\in[1, W]  \nonumber\\    
    L_{local} =  Focal(Up([S_{n, M_l}, S_{a, M_l}]), S) + Dice(Up(S_{n, M_l}), I-S) + Dice(Up(S_{a, M_l}), S), \nonumber
\end{gather}
where $Focal(\cdot, \cdot)$ and $Dice(\cdot, \cdot)$ denote a focal loss~\citep{lin2017focal} and a Dice loss~\citep{li2019dice} respectively. The operators $Up(\cdot)$ and $[\cdot, \cdot]$ represent the unsampling and concatenation along with the channel, and $I$ represents the full-one matrix. Since the anomalous regions are typically smaller than the normal ones, we use focal loss to address the imbalance problem. Furthermore, to ensure that the model establishes an accurate decision boundary, we employ the Dice loss to measure the overlaps between the predicted segmentation $Up(S_{n, M_l})$/$Up(S_{a, M_l})$ and the ground truth mask.

% \vspace{-0.5em}

\paragraph{Refinement of the textual space}
To facilitate the learning of a more discriminative textual space via Eq. \ref{eq:total_loss}, inspired by \cite{jia2022visual} and \cite{khattak2023maple}, we introduce randomly initialized learnable token embeddings in the text encoder to refine the textual space for its adaptation to AD. To control the degree of refinement in the textual space, we choose to replace the prefix part of the original token embeddings with learnable token embeddings in the text encoder, from the bottom (the second layer) to the top. In particular, we denote the token embeddings of the learnable text prompt as \( t_m = \{t^1_m, t^2_m, \dots, t^P_m\} \), which we denote as \( t_m = t^i_m|_1^P \) and \( m \) represents the layer index of the text encoder. We introduce additional multi-layer trainable tokens \( q_m = q^i_m|_1^Q = \{q^1_m, q^2_m, \dots, q^Q_m\} \), \( Q < P \). To adapt the original textual representations of layer \( m \), we replace \(t^i_m|_1^Q\) with \( q_m \), thus deriving the new token embeddings \( t'_m = [q^i_m|_1^Q, t^i_m|_{Q+1}^P] = \{q^1_m, q^2_m, \dots, q^Q_m, t^{Q+1}_m, \dots, t^P_m\} \). Then, \( t'_m \) is forwarded into \( T_m \) to obtain \( t_{m+1} = [r^i_{m+1}|_1^Q, t^i_{m+1}|_{Q+1}^P] =\{r^1_{m+1}, r^2_{m+1}, \dots, r^Q_{m+1}, t^{Q+1}_{m+1}, \dots, t^P_{m+1}\} \). To provide layer-wise adaptation, we discard the obtained \( r^i_{m+1}|_1^Q \) and initialize new learnable token embeddings \( q^i_{m+1}|_1^Q \). Through the concatenation operation, we obtain \( t'_{m+1} = [q^i_{m+1}|_1^Q, t^i_{m+1}|_{Q+1}^P] \), which further refines the textual representations of layer \( m+1 \). We repeat this operation until we reach the designated layer \( M' \). This procedure is given by:  
\begin{align}
    t_2 &= T_1(t_1)  \nonumber\\
    &\ldots \nonumber\\
    [r^i_{m+1}|_1^Q, t^i_{m+1}|_{Q+1}^P] &= T_m([q^i_m|_1^Q, t^i_m|_{Q+1}^P]) \nonumber\\
    [r^i_{m+2}|_1^Q, t^i_{m+2}|_{Q+1}^P] &= T_{m+1}([q^i_{m+1}|_1^Q, t^i_{m+1}|_{Q+1}^P])\\
    &\ldots \nonumber\\
    t_{M'+1} &= T_{M'}([r^i_{M'}|_1^Q, t^i_{M'}|_{Q+1}^P]),\nonumber
\end{align}
where the operator $[\cdot, \cdot]$ represents the concatenation along the channel.

% \vspace{-0.5em}
\paragraph{Refinement of the local visual space}
\label{modifying the visual encoder from the viewpoint of energy}

\begin{wrapfigure}{r}{0.4\textwidth}
\centering
% \vspace{-0.5cm}
   \resizebox{0.4\textwidth}{!}{%
  \subfloat[Input]{\includegraphics[width=0.193\textwidth]{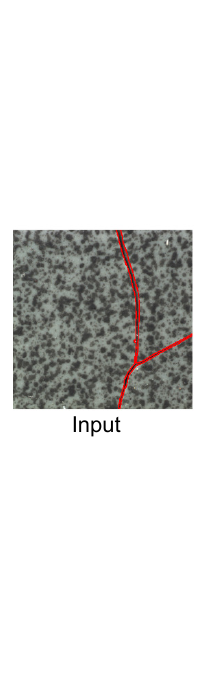}
  \label{input}}
  \hfil
  \subfloat[Q-K attention]{\includegraphics[width=0.193\textwidth]{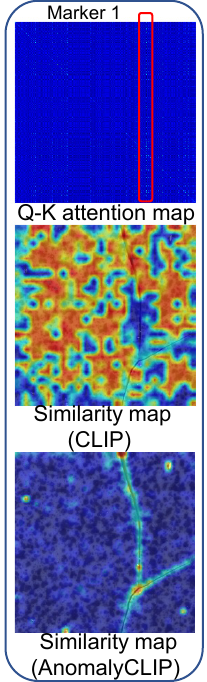}
  \label{q-k-ori}}
  \hfil
  \subfloat[Q-Q attention]{\includegraphics[width=0.193\textwidth]{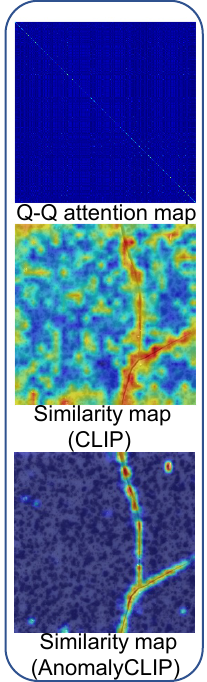}
  \label{q-q}}
  \subfloat[K-K attention]{\includegraphics[width=0.193\textwidth]{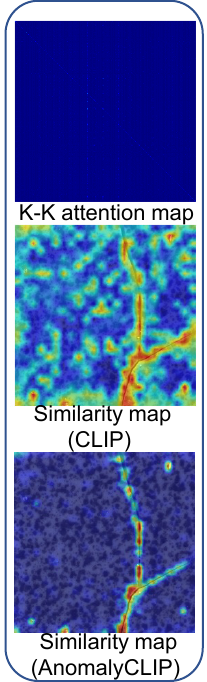}
  \label{k-k}}
  \hfil
  \subfloat[V-V attention]{\includegraphics[width=0.193\textwidth]{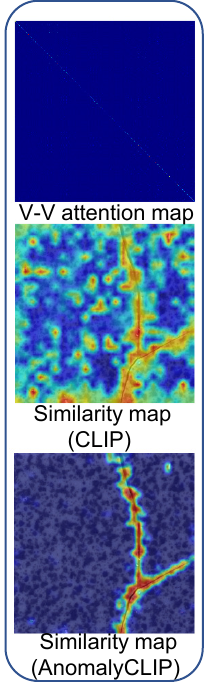}
  \label{v-v}}}

% \setlength{\belowcaptionskip}{-0.1cm}   
% \vspace{-0.3em}
\caption{DPAM visualization.}
% \vspace{-1.2em}
\label{fig6:performance gain between object-agnostic/dependent.}

\end{wrapfigure}

Since the visual encoder of CLIP is originally pre-trained to align global object semantics, the contrastive loss used in CLIP makes the visual encoder produce a representative global embedding for class recognition. Through the self-attention mechanism, the attention map in the visual encoder focuses on the specific tokens highlighted within the red rectangle in Fig~\ref{q-k-ori}. Although these tokens may contribute to global object recognition, they disrupt the local visual semantics, which directly hinders the effective learning of the fine-grained abnormality in our object-agnostic text prompts~\citep{li2023clip}. We found empirically that a diagonally prominent attention map helps reduce the disturbance from other tokens, leading to improved local visual semantics. Therefore, we propose a mechanism called Diagonally Prominent Attention Map to refine the local visual space, with the visual encoder kept frozen during training. To this end, we replace the original $Q$-$K$ attention in the visual encoder with diagonally prominent attention, such as $Q$-$Q$, $K$-$K$, and $V$-$V$ self-attention schemes. As demonstrated in Fig.\ref{q-q}, Fig.\ref{k-k}, and Fig.~\ref{v-v}, the refined DPAM attention maps are more diagonally prominent, resulting in substantially improved segmentation maps in both original CLIP and our AnomalyCLIP. Compared to CLIP that is based on global features and manually defined text prompts, the text prompts learned by AnomalyCLIP are more fine-grained, enabling substantially more accurate alignment between the normality/abnormality prompt embeddings and the local visual embeddings across four different self-attention schemes. This, in turn, allows AnomalyCLIP to generate accurate $S_n$ and $S_a$ for the joint optimization in Eq.~\ref{eq:total_loss}.
Unless otherwise specified, AnomalyCLIP utilizes $V$-$V$ self-attention due to its superior overall performance. The performance of different self-attention mechanisms is analyzed in Sec.~\ref{Additional ablations}. We also provide a detailed explanation about DPAM in Appendix~\ref{Detailed Analysis of DPAM}.

% \vspace{-0.5em}
\paragraph{Training and Inference}
During training, AnomalyCLIP minimizes the loss in Eq.~\ref{eq:total_loss} using an auxiliary AD-related dataset. As for inference, given a test image $x_i$, we use the similarity score $P(g_a, f_i)$ as the image-level anomaly score, with the anomaly score leaning toward one when the anomaly textual embedding $g_a$ is aligned with global visual embedding $f_i$. For pixel-wise predictions, we merge the segmentation $S_{n, M_l}$ and $S_{a, M_l}$ of all selected intermediate layers, followed by an interpolation and smoothing operation. Formally, our anomaly score map $Map \in \R^{H_{image} \times W_{image}}$ is computed as $ Map = G_{\sigma}( \sum_{M_l \in \mathcal{M}}(\frac{1}{2}(I - Up(S_{n, M_l}))+ \frac{1}{2}Up(S_{a, M_l})))$, where $G_{\sigma}$ represents a Gaussian filter, and $\sigma$ controls smoothing.
% \vspace{-0.5em}
\section{Experiments}

\subsection{Experiment setup}

\paragraph{Datasets and Evaluation Metrics}

We conducted extensive experiments on 17 publicly available datasets, covering various industrial inspection scenarios and medical imaging domains (including photography, endoscopy, and radiology) to evaluate the performance of AnomalyCLIP. In industrial inspection, we consider MVTec AD~\citep{bergmann2019mvtec}, VisA~\citep{zou2022spot}, MPDD~\citep{jezek2021deep}, BTAD~\citep{mishra2021vt}, SDD~\citep{tabernik2020segmentation}, DAGM~\citep{wieler2007weakly}, and DTD-Synthetic~\citep{aota2023zero}. In medical imaging, we consider skin cancer detection dataset ISIC~\citep{gutman2016skin}, colon polyp detection datasets CVC-ClinicDB~\citep{bernal2015wm}, CVC-ColonDB~\citep{tajbakhsh2015automated}, Kvasir~\citep{jha2020kvasir}, and Endo~\citep{hicks2021endotect}, thyroid nodule detection dataset TN3k~\citep{gong2021multi}, brain tumor detection datasets HeadCT~\citep{salehi2021multiresolution}, BrainMRI~\citep{salehi2021multiresolution}, Br35H~\citep{br35h}, and COVID-19 detection dataset COVID-19~\citep{9144185, rahman2021exploring}. The SOTA competing methods include CLIP~\citep{radford2021learning}, CLIP-AC~\citep{radford2021learning}, WinCLIP~\citep{jeong2023winclip}, VAND~\citep{chen2023zero}, and CoOp~\citep{zhou2022learning}. We provide more details about the methods and data pre-processing in Appendix~\ref{Implementation details and baselines}. The anomaly detection performance is evaluated using the Area Under the Receiver Operating Characteristic Curve (AUROC). Additionally, average precision (AP) for anomaly detection and AUPRO~\citep{bergmann2020uninformed} for anomaly segmentation are also used to provide more in-depth analysis of the performance.

% \vspace{-0.5em}
\paragraph{Implementation details}
\label{Implementation details}
We use the publicly available CLIP model\footnote{https://github.com/mlfoundations/open\_clip} (\verb'VIT-L/14@336px') as our backbone. Model parameters of CLIP are all frozen. The length of learnable word embeddings $E$ is set to 12. The learnable token embeddings are attached to the first 9 layers of the text encoder for refining the textual space, and $\lambda$ is set to 4. \textcolor{orange}{We use the top feature map as local visual details for anomaly segmentation.} Then, we fine-tune AnomalyCLIP using the test data on MVTec AD and evaluate the ZSAD performance on other datasets. As for MVTec AD, we fine-tune AomalyCLIP on the test data of VisA. We report dataset-level results, which are averaged across their respective sub-datasets. All experiments are conducted in PyTorch-2.0.0 with a single NVIDIA RTX 3090 24GB GPU. More details can be found in Appendix~\ref{Implementation details and baselines}.

\begin{table}[]
\centering
\caption{ZSAD performance comparison on industrial domain. The best performance is highlighted in red, and the second-best is highlighted in blue. $^\dagger$ denotes results taken from original papers.} 
\label{table1:industrial.}
\tiny
\setlength\tabcolsep{5pt} 
\begin{tabular}{cccccccccccc}
\toprule
Task & Category &  Datasets & $|  \mathcal{C} |$ &CLIP & CLIP-AC & WinCLIP & VAND & CoOp  & AnomalyCLIP  \\ \hline
\multirow{7}{*}{\makecell[c]{Image-level \\ (AUROC,  AP)}} & \makecell[c]{Obj \&texture} &MVTec AD &15 & (74.1, 87.6) & (71.5, 86.4) &(\textcolor{red}{91.8}, \textcolor{red}{96.5})$^\dagger$ & (86.1, 93.5)$^\dagger$  &(88.8, 94.8) &(\textcolor{blue}{91.5}, \textcolor{blue}{96.2})  \\

& \multirow{4}{*}{Obj}   & VisA &12  & (66.4, 71.5) & (65.0, 70.1) & (\textcolor{blue}{78.1}, \textcolor{blue}{81.2})$^\dagger$ & (78.0, 81.4)$^\dagger$ &(62.8, 68.1)     & (\textcolor{red}{82.1}, \textcolor{red}{85.4}) \\
&   & MPDD &6  & (54.3, 65.4) & (56.2, 66.0) & (63.6, 69.9) & (\textcolor{blue}{73.0}, \textcolor{blue}{80.2})&(55.1, 64.2)    & (\textcolor{red}{77.0}, \textcolor{red}{82.0})  \\
&   & BTAD &3  & (34.5, 52.5) & (51.0, 62.1) & (68.2, 70.9) & (\textcolor{blue}{73.6}, 68.6)&(66.8, \textcolor{blue}{77.4})    & (\textcolor{red}{88.3}, \textcolor{red}{87.3})\\
&   & SDD  &1  & (65.7, 45.2) & (65.2, 45.7) & (\textcolor{blue}{84.3}, \textcolor{blue}{77.4}) & (79.8, 71.4)&(74.9, 65.1)  & (\textcolor{red}{84.7}, \textcolor{red}{80.0}) \\

&  \multirow{2}{*}{Texture} & DAGM &10 & (79.6, 59.0) & (82.5, 63.7) & (91.8, 79.5) & (\textcolor{blue}{94.4}, \textcolor{blue}{83.8})&  (87.5, 74.6)  & (\textcolor{red}{97.5}, \textcolor{red}{92.3})\\
&   & DTD-Synthetic &12   & (71.6, 85.7) & (66.8, 83.2) & (\textcolor{blue}{93.2}, 92.6) & (86.4, \textcolor{blue}{95.0})&(-, -)  & (\textcolor{red}{93.5}, \textcolor{red}{97.0}) \\ \hline

\multirow{7}{*}{\makecell[c]{Pixel-level \\ (AUROC, PRO)}}   & \makecell[c]{Obj \&texture} &MVTec AD&15 & (38.4, 11.3) & (38.2, 11.6) &(85.1, \textcolor{blue}{64.6})$^\dagger$ & (\textcolor{blue}{87.6}, 44.0)$^\dagger$  &(33.3, 6.7)  & (\textcolor{red}{91.1}, \textcolor{red}{81.4}) \\
& \multirow{4}{*}{Obj}  & VisA &12 & (46.6, 14.8) & (47.8, 17.3) & (79.6, 56.8)$^\dagger$ & (\textcolor{blue}{94.2}, \textcolor{blue}{86.8})$^\dagger$&(24.2, 3.8)& (\textcolor{red}{95.5}, \textcolor{red}{87.0})  \\
& & MPDD &6 & (62.1, 33.0) & (58.7, 29.1) & (76.4, 48.9) & (\textcolor{blue}{94.1}, \textcolor{blue}{83.2})&(15.4, 2.3)  & (\textcolor{red}{96.5}, \textcolor{red}{88.7}) \\
& & BTAD &3 & (30.6, 4.4) & (32.8, 8.3) & (\textcolor{blue}{72.7}, 27.3) & (60.8, \textcolor{blue}{25.0}) &(28.6, 3.8) & (\textcolor{red}{94.2}, \textcolor{red}{74.8}) \\
& & SDD &1 & (39.0, 8.9) & (32.5, 5.8) & (68.8,  24.2) & (\textcolor{blue}{79.8},  \textcolor{blue}{65.1})&(28.9, 7.1)  & (\textcolor{red}{90.6}, \textcolor{red}{67.8}) \\
% \multirow{2}{*}{Texture} 
& \multirow{2}{*}{Texture}& DAGM &10 & (28.2, 2.9) & (32.7, 4.8) & (\textcolor{blue}{87.6}, 65.7) & (82.4, \textcolor{blue}{66.2})&(17.5, 2.1) & (\textcolor{red}{95.6}, \textcolor{red}{91.0}) \\
& & DTD-Synthetic &12 & (33.9, 12.5) & (23.7, 5.5) & (83.9, 57.8) & (\textcolor{blue}{95.3}, \textcolor{blue}{86.9})&(-, -)  & (\textcolor{red}{97.9}, \textcolor{red}{92.3}) \\  
\bottomrule
\end{tabular}%

% \vspace{-1em}
\end{table}

\begin{table}[]
\centering
\caption{ZSAD performance comparison on medical domain. The best performance is highlighted in red, and the second-best is highlighted in blue. Note that the image-level medical AD datasets do not contain segmentation ground truth, so the pixel-level medical AD datasets are different from the image-level datasets.}

\label{table2: Performance comparison (p-AUCROC and p-PRO) on pixel-level anomaly detection.}
\tiny
\setlength\tabcolsep{5pt} 
\begin{tabular}{cccccccccc}
\toprule
Task & Category &Datasets & $|  \mathcal{C} |$ & CLIP & CLIP-AC & WinCLIP & VAND & CoOp & AnomalyCLIP  \\ \hline

\multirow{4}{*}{\makecell[c]{Image-level \\  (AUROC, AP)}} & \multirow{3}{*}{Brain}
   & HeadCT &1 & (56.5, 58.4) & (60.0, 60.7) & (81.8, 80.2) & (\textcolor{blue}{89.1}, \textcolor{blue}{89.4})&(78.4, 78.8)    & (\textcolor{red}{93.4}, \textcolor{red}{91.6}) \\                                                                                                                                                                                                                                                                                       
&  & BrainMRI &1  & (73.9, 81.7) & (80.6, 86.4) & (86.6, \textcolor{blue}{91.5}) & (\textcolor{blue}{89.3}, 90.9)&(61.3, 44.9)  & (\textcolor{red}{90.3}, \textcolor{red}{92.2}) \\
&  & Br35H &1 & (78.4, 78.8) & (82.7, 81.3) & (80.5, 82.2) & (\textcolor{blue}{93.1}, \textcolor{blue}{92.9})&(86.0, 87.5)     & (\textcolor{red}{94.6}, \textcolor{red}{94.7}) \\

& Chest  & COVID-19  &1 & (73.7, 42.4) & (\textcolor{blue}{75.0},  \textcolor{blue}{45.9}) & (66.4, 42.9)&(15.5, 8.5)  & (25.3, 9.2)    & (\textcolor{red}{80.1}, \textcolor{red}{58.7}) \\ \hline

\multirow{6}{*}{\makecell[c]{Pixel-level \\ (AUROC, PRO)}} &  Skin 
& ISIC &1 & (33.1, 5.8) & (36.0, 7.7) & (83.3, 55.1) & (\textcolor{blue}{89.4}, \textcolor{blue}{77.2})&(51.7, 15.9) & (\textcolor{red}{89.7}, \textcolor{red}{78.4}) \\

& \multirow{4 }{*}{Colon}  & CVC-ColonDB &1 & (49.5, 15.8)  & (49.5, 11.5)  & (70.3,32.5) & (\textcolor{blue}{78.4}, \textcolor{blue}{64.6})&(40.5, 2.6)  & (\textcolor{red}{81.9}, \textcolor{red}{71.3})  \\
&  & CVC-ClinicDB &1 & (47.5, 18.9)  & (48.5, 12.6)  & (51.2,13.8) & (\textcolor{blue}{80.5}, \textcolor{blue}{60.7})&(34.8, 2.4)  & (\textcolor{red}{82.9}, \textcolor{red}{67.8})  \\
&  & Kvasir &1 & (44.6, 17.7)  & (45.0, 16.8)  & (69.7, 24.5) & (\textcolor{blue}{75.0}, \textcolor{blue}{36.2})&(44.1, 3.5)  & (\textcolor{red}{78.9}, \textcolor{red}{45.6})  \\
&  & Endo &1 & (45.2, 15.9)  & (46.6, 12.6)  & (68.2, 28.3) & (\textcolor{blue}{81.9}, \textcolor{blue}{54.9})&(40.6, 3.9)  & (\textcolor{red}{84.1}, \textcolor{red}{63.6})  \\

& Thyroid & TN3K &1 & (42.3, 7.3) & (35.6, 5.2) & (70.7, \textcolor{blue}{39.8}) & (\textcolor{blue}{73.6}, 37.8)&(34.0, 9.5)  & (\textcolor{red}{81.5}, \textcolor{red}{50.4}) \\

\bottomrule
\end{tabular}%
% \vspace{-2em}
\end{table}

\subsection{Main results}

\begin{figure}[htbp]
	\begin{minipage}{0.55\textwidth}
    \centering

    % \vspace{-0.5em}
    \includegraphics[width=1\textwidth]{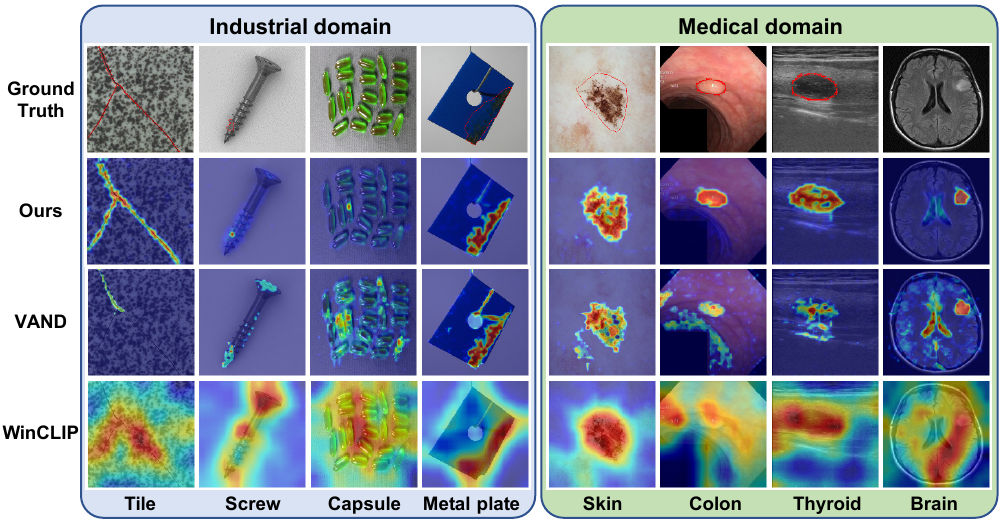}
		\caption{Segmentation visualization.}
		\label{fig4: similarity_score_visualization}
        % \vspace{-1.5em}
	\end{minipage}
	%\qquad
	\hfil
	\begin{minipage}{0.4\textwidth}
		\centering
            \vspace{-0.5em}
            \captionof{table}{ZSAD performance on medical images when fine-tuned by medical image datasets.}%这里必须写table，不然标题就自动设置成figure
           \tiny
           % \vspace{-0.5em}
            \setlength\tabcolsep{1.5pt} 
           \begin{tabular}{cccc}
            \toprule
            Category & Datasets   & VAND  & AnomalyCLIP   \\ \hline
            Classification \\ \hline
            \multirow{3}{*}{Brain}
            & HeadCT & (89.1, 89.4)  & (\textcolor{red}{93.5}, \textcolor{red}{95.1}) \\
            
            % %%%%%%%%%%%%
            & BrainMRI & (89.3,  90.9)  & (\textcolor{red}{95.5}, \textcolor{red}{97.2}) \\
            & Br35H   & (93.1, 92.9) & (\textcolor{red}{97.9}, \textcolor{red}{98.0})  \\
            Chest
            & COVID-19   &(15.5, 8.5) & (\textcolor{red}{70.9}, \textcolor{red}{33.7})\\ \hline
            
            Segmentation \\ \hline
            Skin & ISIC & (58.8, 31.2) & (\textcolor{red}{83.0}, \textcolor{red}{63.8}) \\
            \multirow{3}{*}{Colon} 
            & CVC-ClinicDB  & (89.4, 82.3) & (\textcolor{red}{92.4}, \textcolor{red}{82.9}) \\
            
            & Kvasir & (87.6,  39.3) & (\textcolor{red}{92.5}, \textcolor{red}{61.5)} \\
            & Endo   & (88.5, 81.9)  & (\textcolor{red}{93.2}, \textcolor{red}{84.8)} \\
            Thyroid
            & TN3K & (60.5, 16.8)  & (\textcolor{red}{79.2}, \textcolor{red}{47.0})\\ \hline
            
            \end{tabular}%
            \label{table3: Performance comparison (p-AUCROC and p-PRO) on pixel-level anomaly detection.}
            \vspace{-1.5em}
	    \end{minipage}
\end{figure}

\paragraph{ZSAD performance on diverse industrial inspection domains}
Table~\ref{table1:industrial.} shows the ZSAD results of AnomalyCLIP with five competing methods over seven industrial defect datasets of very different foreground objects, background, and/or anomaly types. AnomalyCLIP achieves superior ZSAD performance across the datasets, substantially outperforming the other five methods in most datasets. The weak performance of CLIP and CLIP-AC can be attributed to CLIP's original pre-training, which focuses on aligning object semantics rather than anomaly semantics. By using manually defined text prompts, WinCLIP and VAND achieve better results. Alternatively, CoOp adopts learnable prompts to learn the global anomaly semantics. However, those prompts focus on the global feature and ignore the fine-grained local anomaly semantics, leading to their poor performance on anomaly segmentation. To adapt CLIP to ZSAD, AnomalyCLIP learns object-agnostic text prompts to focus on learning the generic abnormality/normality using global and local context optimization, enabling the modeling of both global and local abnormality/normality. Our resulting prompts can also generalize to different datasets from various domains. To provide more intuitive results, we visualize the anomaly segmentation results of AnomalyCLIP, VAND, and WinCLIP across different datasets in Fig.~\ref{fig4: similarity_score_visualization}. Compared to VAND and WinCLIP, AnomalyCLIP can perform much more accurate segmentation for the defects from different industrial inspection domains. 

% \vspace{-0.5em}
\paragraph{Generalization from defect datasets to diverse medical domain datasets}  
To evaluate the generalization ability of our model, we further examine the ZSAD performance of AnomalyCLIP on 10 medical image datasets of different organs across different imaging devices.  Table~\ref{table2: Performance comparison (p-AUCROC and p-PRO) on pixel-level anomaly detection.} shows the results, where learning-based methods, including AnomalyCLIP, VAND and CoOp, are all tuned using MVTec AD data. It is remarkable that methods like AnomalyCLIP and VAND obtain promising ZSAD performance on various medical image datasets, even though they are tuned using a defect detection dataset. Among all these methods, AnomalyCLIP is the best performer due to its strong generalization brought by object-agnostic prompt learning. As illustrated in Fig.~\ref{fig4: similarity_score_visualization}, AnomalyCLIP can accurately detect various types of anomalies in diverse medical images, such as skin cancer regions in photography images, colon polyps in endoscopy images, thyroid nodules in ultrasound images, and brain tumors in MRI images, having substantially better performance in locating the abnormal lesion/tumor regions than the other two methods WinCLIP and VAND. This again demonstrates the superior ZSAD performance of AnomalyCLIP in datasets of highly diverse object semantics from medical imaging domains.

\paragraph{Can we obtain better ZSAD performance if fine-tuned using medical image data?}
Comparing the promising performance in industrial datasets, AnomalyCLIP presents a relatively low performance in medical datasets. This is partly due to the impact of auxiliary data used in our prompt learning. So, then we examine whether the ZSAD performance on medical images can be improved if the prompt learning is trained on an auxiliary medical dataset. One challenge is that there are no available large 2D medical datasets that include both image-level and pixel-level annotations for our training. To address this issue, we create such a dataset based on ColonDB (More details see Appendix~\ref{Implementation details and baselines}), and then optimize the prompts in AnomalyCLIP and VAND using this dataset and evaluate their performance on the medical image datasets. 
% We compare the results of VAND and AnomalyCLIP because CLIP and WinCLIP do not use auxiliary data. As 
The results are presented in Table~\ref{table3: Performance comparison (p-AUCROC and p-PRO) on pixel-level anomaly detection.}. AnomalyCLIP and VAND largely improve their detection and segmentation performance compared to that fine-tuned on MVTec AD, especially for the colon polyp-related datasets such as CVC-ClincDB, Kvasir, and Endo (note that these datasets are all from different domains compared to the fine-tuning ColonDB dataset). AnomalyCLIP also exhibits performance improvement in detecting brain tumors in datasets such as HeadCT, BrainMRI, and Br35H. This is attributed to the visual similarities between colon polyps and brain tumors. Conversely, the symptom of the colon polyp differs significantly from that of diseased skin or chest, leading to performance degradation in ISIC and COVID-19. Overall, compared to VAND, AnomalyCLIP performs consistently better across all datasets of anomaly detection and segmentation.

\newpage

\begin{wrapfigure}{r}{0.6\textwidth}
    \centering
    % \vspace{-0.5cm}
    \includegraphics[width=0.6\textwidth]{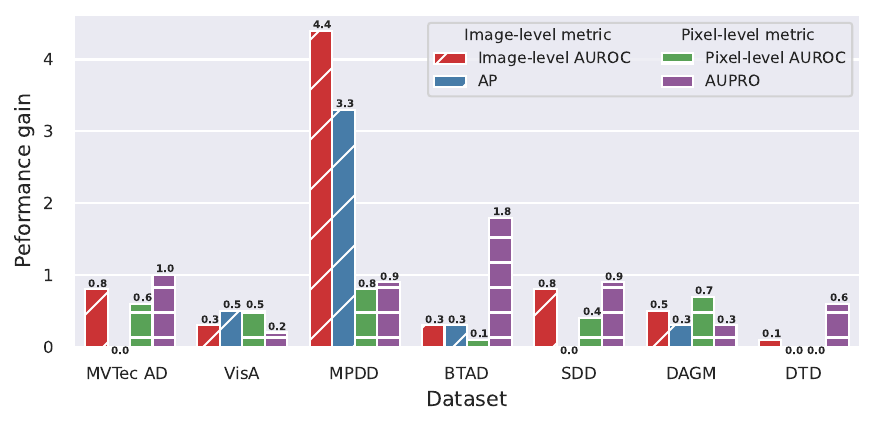}
        \setlength{\abovecaptionskip}{-0.1cm}    
    % \setlength{\belowcaptionskip}{-0.1cm}   
    % \vspace{-1em}
    \caption{Performance gain of using object-agnostic prompts compared to object-aware prompts.}
    % in AnomalyCLIP.}
    % \vspace{-1.2em}
    \label{fig6:performance gain between object-agnostic/aware.}
\end{wrapfigure}
\paragraph{Object-agnostic vs. object-aware prompt learning} To study the effectiveness of object-agnostic prompt learning in AnomalyCLIP, we compare AnomalyCLIP with its variant that uses an object-aware prompt template. The performance gain of AnomalyCLIP to its object-aware prompt learning variant is shown in Fig.~\ref{fig6:performance gain between object-agnostic/aware.}, where positive values indicate our object-agnostic prompt templates are better than the object-aware one. It is clear that our object-agnostic prompt learning performs much better than, or on par with, the object-aware version in both image-level and pixel-level anomaly detection. This indicates that having object-agnostic prompts helps better learn the generic abnormality and normality in images, as the object semantics are often not helpful, or can even become noisy features, for the ZSAD task.

\subsection{Ablation study}
\label{Ablation study}

\paragraph{Module ablation} We first validate the effectiveness of different high-level modules of our AnomalyCLIP, including DPAM ($T_1$), object-agnostic text prompts ($T_2$), and added learnable tokens in text encoders ($T_3$). As shown in Table \ref{table-operators}, each module contributes to the remarkable performance of AnomalyCLIP. DPAM improves the segmentation performance by enhancing local visual semantics ($T_1$). Object-agnostic text prompts focus on the abnormality/normality within images instead of the object semantics, allowing AnomalyCLIP to detect anomalies in diverse unseen objects. Therefore, introducing object-agnostic text prompts ($T_2$) significantly improves AnomalyCLIP. Furthermore, text prompt tuning ($T_3$) also brings performance improvement via the refinement of original textual space.

\begin{figure*}[]

  \begin{minipage}{0.48\textwidth}
        \centering
    
\vspace{-1em}
 \captionof{table}{Module ablation.}
 \vspace{-1em}
 \label{table-operators}
 \tiny
    \setlength\tabcolsep{3pt} 
    \begin{tabular}{ccccc}
    \toprule
    
    % \multicolumn{5}{c|}{Proposed technology}& \multicolumn{4}{c}{MVTEC}\\ \hline
    \multirow{2}{*}{Module} & \multicolumn{2}{c}{MVTec AD} & \multicolumn{2}{c}{VisA} \\ 
    & Pixel-level  & Image-level& Pixel-level  & Image-level   \\ \hline
     Base &(46.8, 15.4)&(66.3, 83.3) &(47.9, 17.1)&(54.4, 61.7)               \\
     $+T_1$ &(68.4, 47.4)& (66.3, 83.3) &(54.8, 32.7)& (54.4, 61.7)       \\   
     $+T_2$ &(89.5, \textcolor{blue}{81.2})& (\textcolor{blue}{90.8}, \textcolor{blue}{96.0}) &(\textcolor{blue}{95.0}, \textcolor{blue}{85.3})& (\textcolor{blue}{81.7}, \textcolor{blue}{85.2})  \\
     $+T_3$ &(\textcolor{red}{91.1}, \textcolor{red}{81.4})& (\textcolor{red}{91.5}, \textcolor{red}{96.2}) &(\textcolor{red}{95.5}, \textcolor{red}{87.0})& (\textcolor{red}{82.1}, \textcolor{red}{85.4})   \\
    
    \bottomrule
    \end{tabular}%
    \end{minipage} 
    \hfill
    \hfill
\begin{minipage}{0.48\textwidth}
    \centering

    \vspace{-1em}
    \captionof{table}{Context optimization ablation.}%这里必须写table，不然标题就自动设置成figure
        \vspace{-1em}
    \tiny
 \label{Context optimization ablation.}
    \setlength\tabcolsep{3pt} 
    \begin{tabular}{cc|cccc}
    \toprule
    
    % \multicolumn{5}{c|}{Proposed technology}& \multicolumn{4}{c}{MVTEC}\\ \hline
    \multirow{2}{*}{Local.} & \multirow{2}{*}{Global.} & \multicolumn{2}{c}{MVTec AD} & \multicolumn{2}{c}{VisA} \\ 
    &  & Pixel-level  & Image-level& Pixel-level  & Image-level   \\ \hline
    \ding{55} & \ding{55} &(71.7, 57.7)&(\textcolor{orange}{68.8}, \textcolor{orange}{85.8}) &(\textcolor{orange}{74.7}, \textcolor{orange}{62.1})&(\textcolor{orange}{61.1}, \textcolor{orange}{69.1}) \\
    \ding{55} & \ding{51} &(80.3, 77.8)& (\textcolor{blue}{89.9}, 95.4) & (86.6, 78.1)& (\textcolor{red}{82.2}, \textcolor{blue}{84.9}) \\ 
    \ding{51} & \ding{55} &(\textcolor{blue}{91.0}, \textcolor{blue}{80.4})& (\textcolor{blue}{89.9}, \textcolor{blue}{96.0}) &(\textcolor{blue}{95.2}, \textcolor{blue}{86.5})& (79.5, 83.2)   \\
    \ding{51} & \ding{51} &(\textcolor{red}{91.1}, \textcolor{red}{81.4})& (\textcolor{red}{91.5}, \textcolor{red}{96.2}) &(\textcolor{red}{95.5}, \textcolor{red}{87.0})& (\textcolor{blue}{82.1}, \textcolor{red}{85.4})   \\
    \bottomrule
    \end{tabular}%

	\end{minipage}
        \vspace{-1em}
\end{figure*}

\begin{figure*}[]
\centering
 \includegraphics[width=0.8\textwidth]{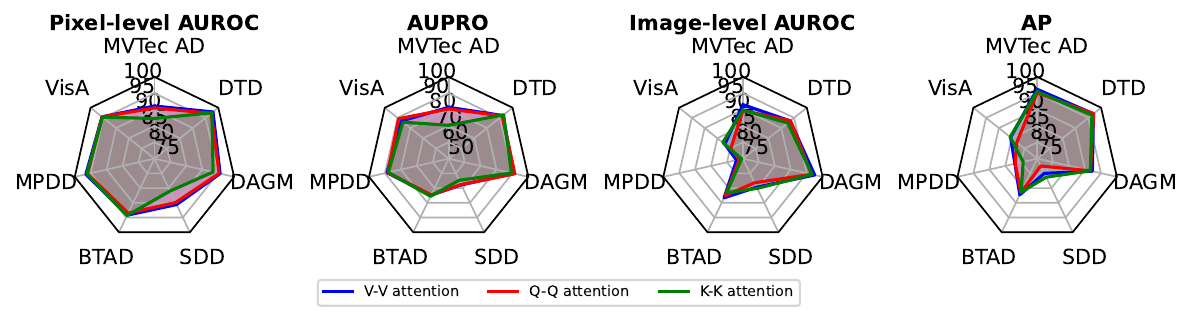}

    \caption{DPAM component ablation.}
    % \vspace{-1.2em}
\label{fig8: DPAM component ablation.}
\end{figure*}

\vspace{-0.5em}
\paragraph{Context optimization}
Next we examine key modules in detail. The object-agnostic prompt learning is the most effective module, and it is driven by our glocal context optimization, so we consider two different optimization terms, local and global losses, in Eq.~\ref{eq:total_loss}. The results are shown in Table~\ref{Context optimization ablation.}.
% , we conduct experiments involving global context optimization and local context optimization separately. 
Both global and local context optimization contribute to the superiority of AnomalyCLIP. Global context optimization helps to capture global anomaly semantics, thus enabling more accurate image-level detection. Compared to global context optimization, local context optimization incorporates local anomaly semantics, which improves pixel-level performance and complements image-level performance. 
% Meanwhile, it also promotes the alignment of the global feature. 
By synthesizing these two optimization strategies, AnomalyCLIP generally achieves better performance than using them individually.

\vspace{-0.5em}
\paragraph{DPAM strategy ablation}
AnomalyCLIP uses $V$-$V$ self-attention by default. Here we study the effectiveness of using two other DPAM strategies, including $Q$-$Q$ and $K$-$K$ self-attention, resulting in two AnomalyCLIP variants, namely AnomalyCLIP$_{qq}$ and AnomalyCLIP$_{kk}$. The comparison results are presented in Fig.~\ref{fig8: DPAM component ablation.}. AnomalyCLIP$_{qq}$ achieves similar segmentation capabilities as AnomalyCLIP but suffers from degradation in detecting image-level anomalies. Conversely, while AnomalyCLIP$_{kk}$ performs well in anomaly classification, its segmentation performance is less effective than AnomalyCLIP and AnomalyCLIP$_{qq}$. The $V$-$V$ self-attention is generally recommended in AnomalyCLIP. \textcolor{orange}{Detailed analysis of DPAM can be seen in Appendix~\ref{Detailed Analysis of DPAM}.}
\vspace{-0.5em}

% \vspace{-0.8em}
\section{Related Work}
\vspace{-0.5em}
% \subsection{Zero-shot anomaly detection}
\paragraph{Zero-shot anomaly detection}
ZSAD relies on the model's strong transferability to handle unseen anomalies\textcolor{orange}{~\citep{ aota2023zero}}. 
\textcolor{orange}{CLIP-AD~\citep{liznerski2022exposing} and ZOC~\citep{esmaeilpour2022zero} are early studies in utilizing CLIP for ZSAD, but they mainly focus on the anomaly classification task. ACR~\citep{li2023zero} requires tuning on target-domain-relevant auxiliary data for ZSAD on different target datasets, while AnomalyCLIP can be applied to different datasets after it is trained on one general dataset.} A very recent approach WinCLIP~\citep{jeong2023winclip} presents a seminal work that leverages CLIP for zero-shot classification and segmentation. It uses a large number of hand-crafted text prompts and involves multiple forward passes of image patches for anomaly segmentation. To tackle this inefficiency, VAND~\citep{chen2023zero} introduces learnable linear projection techniques to enhance the modeling of local visual semantics. However, these approaches suffer from insufficiently generalized textual prompt embeddings, which degrades their performance in identifying anomalies associated with various unseen object semantics. AnomalyCLIP utilizes only two object-agnostic learnable text prompts to optimize the generic text prompts of abnormality and normality, and it can obtain segmentation results with just a single forward pass. \textcolor{orange}{AnomalyGPT~\citep{gu2023anomalygpt} is a concurrent work in utilizing foundation models for AD, but it is designed for unsupervised/few-shot AD with manually crafted prompts.}
% \vspace{-1em}
% \subsection{Prompt learning}
\paragraph{Prompt learning}
Rather than resorting to full network fine-tuning, prompt learning emerges as a parameter-efficient alternative to achieve satisfactory results~\citep{sun2022dualcoop, khattak2023maple, kim2023zegot, zhou2022conditional}. CoOp~\citep{zhou2022learning} introduces learnable text prompts for few-shot classification. On this basis, DenseCLIP~\citep{rao2022denseclip} extends prompt learning to dense prediction tasks with an extra image decoder. Instead, AnomalyCLIP proposes object-agnostic prompt learning for anomaly detection, blocking out the potential adverse impact of the diverse object semantics on anomaly detection. Benefiting from the glocal context optimization, AnomalyCLIP can capture local anomaly semantics such that we can simultaneously perform classification and segmentation tasks without an additional decoder network like~\cite{rao2022denseclip}.

\vspace{-0.8em}
\section{Conclusion}
In this paper, we tackle a challenging yet significant area of anomaly detection, ZSAD, in which there is no available data in the target dataset for training. We propose AnomalyCLIP to improve the weak generalization performance of CLIP for ZSAD. We introduce object-agnostic prompt learning to learn generic abnormality/normality text prompts for generalized ZSAD on image datasets of diverse foreground objects. Further, to incorporate global and local anomaly semantics into AnomalyCLIP, we devise a joint global and local context optimization to optimize the object-agnostic text prompts. Extensive experimental results on 17 public datasets demonstrate that AnomalyCLIP achieves superior ZSAD performance.

\subsubsection*{Acknowledgments}
This work was supported by NSFC U1909207, NSFC 62088101 Autonomous Intelligent Unmanned Systems, and the Singapore Ministry of Education Academic Research Fund Tier 1 grant (21SISSMU031). 
% (In this work Q. Zhou, S. He, and J. Chen were supported by NSFC under grant No. U1909207).

\subsubsection*{Reproducibility Statement}

To ensure the reproducibility and completeness of this paper, we have included an Appendix consisting of five main sections. In Appendix~\ref{Implementation details and baselines}, we provide more implementation details of AnomalyCLIP, as well as the reproduction of other baseline methods. Appendix~\ref{Dataset} provides key statistics about the datasets used in our experiments and the implementation of the auxiliary medical dataset for prompt tuning. Appendix~\ref{Additional ablations} supplements the main paper with additional results and ablations. Further visualizations of similarity scores and maps are detailed in Appendix~\ref{Visualization}. Additionally, the main paper presents only the average performance in each dataset that contains a number of data subsets, for which we present their fine-grained detection results, in Appendix~\ref{Class-level performance}. Our code will be made publicly accessible once the paper is accepted.

\bibliography{iclr2024_conference}
\bibliographystyle{iclr2024_conference}

\appendix

\section{Implementation details and baselines}
\label{Implementation details and baselines}
\subsection{Implementation details}In this paper, we use the publicly available CLIP model (\verb'VIT-L/14@336px') as our backbone. Model parameters of CLIP are all frozen. The length of learnable text prompts $M$ is set to 12. These trainable text tokens are attached to the first 9 layers of the text encoder, and each text token has a length of 4. We fine-tune AnomalyCLIP on the test data on MVTec AD and test the performance for other datasets. As for MVTec AD, we fine-tune AomalyCLIP on test data on VisA. To provide adequate visual details, we extract the local feature map from the top layer of the visual encoder. Starting from the 6th layer, we apply DPAM to the architecture of the visual encoder according to Sec.~\ref{modifying the visual encoder from the viewpoint of energy}. Additionally, we set the balanced weight $\lambda$ to 4 in our loss function. The input images are resized to a size of 518 with batch size 8, and we use the Adam optimizer~\citep{kingma2014adam} with a learning rate of 0.001 to update model parameters.  During testing, we apply a Gaussian filter with $\sigma=4$ to smooth the anomaly score map. The epoch is 15 for all experiments, which are performed in PyTorch-2.0.0 with a single NVIDIA RTX 3090 24GB GPU.

\subsection{Baselines}
To demonstrate the superiority of Anomlay-CLIP, we compare AnomlayCLIP with broad SOTA baselines.
Implementation and reproduction details are given as follows:
\begin{itemize}
    \item CLIP~\citep{radford2021learning}. CLIP is a powerful zero-shot classification method. To perform the anomaly detection task, we use two classes of text prompt templates \verb'A photo of a normal [cls]' and \verb'A photo of an anomalous [cls]', where \verb'cls' denotes the target class name. The anomaly score is computed according to Eq.~\ref{equ: softmax}. As for anomaly segmentation, we extend the above computation to local visual embedding to derive the segmentation.  
    \item CLIP-AC~\citep{radford2021learning}. Different from CLIP, CLIP-AC employs an ensemble of text prompt templates that are recommended for ImageNet dataset~\citep{radford2021learning}. We average the generated textual embeddings of normal and anomaly classes respectively, and compute the probability and segmentation in the same way as CLIP.
    \item WinCLIP~\citep{jeong2023winclip}. WinCLIP is a SOTA ZSAD method. They design a large set of hand-crafted text prompt templates specific to anomaly detection and use a window scaling strategy to obtain anomaly segmentation. All parameters are kept the same as in their paper.
    \item VAND~\citep{chen2023zero}. VAND is an improved version of WinCLIP. They first adjust the text prompt templates and then introduce learnable linear projections to improve local visual semantics to derive more accurate segmentation. All parameters are kept the same as in their paper.
    \item CoOp~\citep{zhou2022learning}. CoOp is a representative method for prompt learning. To adapt CoOp to ZSAD, we replace its learnable text prompt templates $[V_1][V_2]...[V_N][\verb'cls']$ with normality and abnormality text prompt templates, where $V_i$ is the learnable word embeddings. The normality text prompt template is defined as $[V_1][V_2]...[V_N][\verb'normal'][\verb'cls']$, and the abnormality one is defined as $[V_1][V_2]...[V_N][\verb'anomalous'][\verb'cls']$. Anomaly probabilities and segmentation are obtained in the same way as for AnomalyCLIP. All parameters are kept the same as in their paper.
    \end{itemize}

\section{Dataset}
\label{Dataset}
\begin{table}[]
\caption{Key statistics on the datasets used.}
\label{table: Statistical information on the Datasets.}
\resizebox{1\textwidth}{!}{%
\begin{tabular}{cccccc}
\toprule
Dataset &Category &Modalities & $|  \mathcal{C} |$  & \makecell[c]{Normal and \\anomalous samples} &Usage \\
\hline
\href{https://www.mvtec.com/company/research/datasets/mvtec-ad}{MVTec AD}& \makecell[c]{Obj \&texture}&Photography& 15 & (467, 1258) &Industrial defect detection \\ \hline
\href{https://github.com/amazon-science/spot-diff}{VisA}& \multirow{4}{*}{Obj} &Photography & 12 & (962, 1200) &Industrial defect detection  \\
\href{https://github.com/amazon-science/spot-diff}{MPDD}&    &Photography & 6 &(176, 282) &Industrial defect detection  \\
\href{http://avires.dimi.uniud.it/papers/btad/btad.zip}{BTAD}&    &Photography & 3& (451, 290) &Industrial defect detection  \\
\href{https://www.vicos.si/resources/kolektorsdd/}{SDD}&    &Photography &1 & (181, 74) &Industrial defect detection  \\ \hline
\href{https://www.kaggle.com/datasets/mhskjelvareid/dagm-2007-competition-dataset-optical-inspection}{DAGM}& \multirow{2}{*}{Texture}&Photography & 10 & (6996, 1054) &Industrial defect detection  \\
\href{https://drive.google.com/drive/folders/10OyPzvI3H6llCZBxKxFlKWt1Pw1tkMK1}{DTD-Synthetic}& &Photography & 12 & (357, 947) &Industrial defect detection  \\\hline
\href{https://isic-challenge-data.s3.amazonaws.com/2016/ISBI2016_ISIC_Part1_Test_Data.zip}{ISIC}&Skin  &Photography  & 1 & (0, 379) &Skin cancer detection  \\\hline

\href{https://figshare.com/articles/figure/Polyp_DataSet_zip/21221579}{CVC-ClinicDB} &   & Endoscopy& 1 & (0, 612) &Colon polyp detection \\
\href{https://figshare.com/articles/figure/Polyp_DataSet_zip/21221579}{CVC-ColonDB} & &Endoscopy & 1 & (0, 380) &Colon polyp detection \\
\href{https://figshare.com/articles/figure/Polyp_DataSet_zip/21221579}{Kvasir} &  & Endoscopy & 1 & (0, 1000) &Colon polyp detection \\
\href{https://drive.google.com/file/d/1LNpLkv5ZlEUzr_RPN5rdOHaqk0SkZa3m/view}{Endo} &  &Endoscopy & 1 & (0, 200) &Colon polyp detection \\\hline
\href{https://github.com/haifangong/TRFE-Net-for-thyroid-nodule-segmentation?tab=readme-ov-file}{TN3K} &Thyroid&\makecell[c]{Radiology \\(Utralsound)}& 1 & (0, 614) &Thyroid nodule detection\\\hline
\href{https://www.kaggle.com/datasets/felipekitamura/head-ct-hemorrhage}{HeadCT} &\multirow{6}{*}{Brain} &\makecell[c]{Radiology \\(CT)} &1 & (100, 100) &Brain tumor detection \\
\href{https://www.kaggle.com/datasets/navoneel/brain-mri-images-for-brain-tumor-detection}{BrainMRI} & &\makecell[c]{Radiology \\(MRI)} &1 & (98, 155) &Brain tumor detection \\
\href{https://www.kaggle.com/datasets/ahmedhamada0/brain-tumor-detection}{Br35H} & &\makecell[c]{Radiology \\(MRI)} &1 & (1500, 1500) &Brain tumor detection \\ \hline
\href{https://www.kaggle.com/datasets/tawsifurrahman/covid19-radiography-database}{COVID-19} &Chest &\makecell[c]{Radiology \\(X-ray)}&1 & (1341, 219) &COVID-19 detection \\
\bottomrule
\end{tabular}%
}
\end{table}

\paragraph{More dataset details}
In this paper, we conduct extensive experiments on 17 public datasets spanning two domains and three modalities to validate the effectiveness of our methods. Since we just use the test data of Datasets, we present the relevant information of their test sets in Table~\ref{table: Statistical information on the Datasets.}. 
We apply the default normalization of OpenCLIP to all datasets. After normalization, we resize the images to a resolution of (518, 518) to obtain an appropriate visual feature map resolution. It should be noted that the original image size of SDD has a width of 500 and a height ranging from 1,240 to 1,270. Before processing, we vertically divide the original 500 × 1,250 image into two images and assign pixel-wise annotations to each image.

\paragraph{Fine-tuning medical dataset}
We cannot find publicly available 2D medical AD datasets that include both category labels and segmentation ground truths simultaneously. To fill the blank, in this paper, we create such a medical dataset by combining two existing 2D medical datasets. Particularly, we use the colon polyp detection dataset ColonDB~\citep{tajbakhsh2015automated} to provide pixel-level annotations. Meanwhile, considering the normal samples in the same domain, we choose the test split of Endo classification dataset~\citep{hicks2021endotect} to combine with ColonDB. As a result, the new medical dataset contains 163 normal samples and 380 anomaly samples, supporting both anomaly classification and segmentation tasks.

\textcolor{orange}{
\section{Detailed Analysis of DPAM}
\label{Detailed Analysis of DPAM}
Since the visual encoder of CLIP is originally pre-trained to align
global object semantics, such as cat and dog, the contrastive loss used in CLIP makes the visual encoder produce a representative global embedding for recognizing semantic classes. Through the self-attention mechanism, the attention map in the visual encoder focuses on the specific tokens highlighted within the red rectangle in Fig.~\ref{q-k-ori}. Although these tokens may contribute to global object recognition, they disrupt the local visual semantics, which directly hinders the effective learning of the fine-grained abnormality in our object-agnostic text prompts. For segmentation purposes, it's crucial for the visual feature map to emphasize the surrounding context to capture more local visual semantics.\\
Formally, let $a_{ij}$ be an attention score in the attention score matrix, where $i, j \in [1, h \times w]$, then the i-th output of $Q$-$K$ attention can be written as: 
\begin{equation}Attention( {Q}, {K}, {V})_i =  softmax\left(\frac{q_iK^{\top}}{\sqrt{D}}\right)V =\frac{\sum\limits_{j=1}^n a_{ij} v_j}{\sum\limits_{j=1}^n a_{ij}}\label{eq:std-att-2}, \quad \quad a_{ij} =  e^{  \frac{q_i k_j^{\top}}{\sqrt{D}}}. \nonumber
\end{equation}
Note that vectors (i.e., $q_i$, $k_i$, $v_i$) are represented as row vectors. $Attention( {Q}, {K}, {V})_i$ can be regarded as the weighted average of $v_j$ using $a_{ij}$ as the weight. Assuming that the original attention map focuses on the specific tokens at index $m$, it is clear that $q_i$ only produces the large attention score with $k_m$ in all $k_j$. Therefore, $a_{im}$ is the largest score among other $a_{ij}$ so $Attention({Q}, {K}, {V})_i$ is dominated by $v_m$, which causes the local visual embedding at index $i$ to be disturbed by the local visual embedding at index $m$. In Figure 3(b), the attention score map presents vertical activation and suggests that every $q_i$ produces a large attention score with $k_m$. In such a case, several $Attention({Q}, {K}, {V})_i$ is dominated by $v_m$ and results in weak anomaly segmentation in Figure 3(b) even though $v_m$ may be important for original class recognition. Some prior studies~\citep{rao2022denseclip, gu2023anomalygpt} use an additional decoder to recover the local visual semantics. In this paper, we directly use local visual embeddings for segmentation and point out that an ideal attention map for local visual semantics should exhibit a more pronounced diagonal pattern. For this purpose, DPAM is proposed to replace the original $Q$-$K$ attention with analogous components, including $Q$-$Q$, $K$-$K$, and $V$-$V$ self-attention. Therefore, $a_{ij}$ is changed into:
\begin{equation}
    a^{qq}_{ij} =  e^{ \frac{q_i q_j^{\top}}{\sqrt{D}}}, \quad  a^{kk}_{ij} =  e^{ \frac{k_i k_j^{\top}}{\sqrt{D}}}, \quad a^{vv}_{ij} =  e^{ \frac{v_i v_j^{\top}}{\sqrt{D}}}.  \nonumber
\end{equation}
This modification ensures that $q_i$, $k_i$, and $v_i$ hold significant weight in forming $Attention({Q}, {Q}, {V})_i$, $Attention({K}, {K}, {V})_i$, and $Attention({V}, {V}, {V})_i$, thereby preserving local visual semantics. As a result, the produced attention maps exhibit a more diagonal prominence compared to the original Q-K attention, leading to improved performance in anomaly segmentation, as shown in Fig.\ref{q-q}, Fig.\ref{k-k}, and Fig.~\ref{v-v}. However, since $Q$ and $K$ consist of the original attention map, other important tokens at index $n$ for class recognition within themselves may also produce relatively large scores ($a_{in}$) (e.g., $q_i$ has strong relevance with $q_n$  
besides $q_i$) to disturb $Attention({Q}, {Q}, {V})_i$ and $Attention({K}, {K}, {V})_i$ Fig.\ref{q-q} and Fig.\ref{k-k}. In contrast to $Q$-$Q$ and $K$-$K$, $V$-$V$ does not participate in computing the original attention map, reducing the unexpected bias to different tokens in $V$ for the purpose of anomaly segmentation. Therefore, $v_i$ does not produce a large weight ($a_{ij}$) with $v_j$ and generates a larger weight ($a_{ii}$) to form $Attention({V}, {V}, {V})_i$, preserving more information of $v_i$ and experiencing diagonally prominent attention map (minimal disturbance), as depicted in Fig.~\ref{v-v}. This is the reason why $V$-$V$ achieves the best results.
}

\section{Additional results and ablations}
\label{Additional ablations}

%%%%%%%%%%%%%%%%%%%%%%%%%%%%%%%%%%%%%%%%%%%%%%%%%%%%%%%%%%%%%%%%%%%%%%%%%%%%%%%%%%%%%%%%%%%%%%%%%%%%%
\begin{table}[h]
\centering
\caption{Comparison of ZSAD performance between AnomalyCLIP and SOTA full-shot methods. The best performance is highlighted in red, and the second-best is highlighted in blue.} 
\label{table1: Performance comparison (p-AUCROC and p-PRO) on pixel-level anomaly detection.}
\tiny

\begin{tabular}{cccccccc}
\toprule
Task & Category &  Datasets & $|  \mathcal{C} |$  & AnomalyCLIP  &PatchCore & RD4AD\\ \hline
\multirow{6}{*}{\makecell[c]{Image-level \\ (AUROC, AP)}} & \makecell[c]{Obj \&texture} & MVTec AD & 15  &(91.5, 96.2) &(\textcolor{red}{99.0}, \textcolor{red}{99.7}) & \textcolor{blue}{(98.7}, \textcolor{blue}{99.4}) \\

& \multirow{4}{*}{Obj}   & VisA & 12    & (82.1, 85.4) & (\textcolor{red}{94.6}, \textcolor{red}{95.9}) & (\textcolor{red}{95.3}, \textcolor{blue}{95.7})\\
&   & MPDD & 6 &  (77.0, 82.0) & (\textcolor{red}{94.1}, \textcolor{red}{96.3}) & (\textcolor{blue}{91.6}, \textcolor{blue}{93.8})\\
&   & BTAD  & 3   & (88.3, 87.3) & (\textcolor{blue}{93.2}, \textcolor{red}{98.6}) & (\textcolor{red}{93.8}, \textcolor{blue}{96.8})\\
&   & SDD & 1   & (\textcolor{blue}{84.7}, \textcolor{blue}{80.0}) & (64.9, 48.3)  & (\textcolor{red}{86.8}, \textcolor{red}{81.3})\\

&  \multirow{1}{*}{Texture} & DAGM  & 10  & (\textcolor{red}{97.5}, \textcolor{red}{92.3}) & (92.7, \textcolor{blue}{81.3}) & (\textcolor{blue}{92.9}, 79.1)\\ \hline

\multirow{6}{*}{\makecell[c]{Pixel-level \\ (AUROC, PRO)}}   & \makecell[c]{Obj \&texture} &MVTec AD & 15  &(91.1, 81.4) & (\textcolor{red}{98.1}, \textcolor{blue}{92.8}) & (\textcolor{blue}{97.8}, \textcolor{red}{93.6})\\
& \multirow{4}{*}{Obj}  & VisA & 12  & (95.5, 87.0)  & (\textcolor{red}{98.5}, \textcolor{red}{92.2}) & (\textcolor{blue}{98.4}, \textcolor{blue}{91.2})\\
& & MPDD & 6 & (96.5, 88.7) & (\textcolor{red}{98.8}, \textcolor{blue}{94.9}) & (\textcolor{blue}{98.4}, \textcolor{red}{95.2})\\
& & BTAD & 3  & (94.2, \textcolor{blue}{74.8}) & (\textcolor{blue}{97.4}, 74.4) &(\textcolor{red}{97.5}, \textcolor{red}{75.1})\\
& & SDD & 1  & (\textcolor{blue}{90.6}, \textcolor{blue}{67.8}) & (87.9, 46.3) &(\textcolor{red}{92.2}, \textcolor{red}{72.0})\\
% \multirow{2}{*}{Texture} 

& \multirow{1}{*}{Texture}& DAGM & 10 & (95.6, \textcolor{blue}{91.0})  &(\textcolor{blue}{95.9}, 87.9)&  (\textcolor{red}{96.8}, \textcolor{red}{91.9})\\
% & & DTD-Synthetic  & (97.9, 92.3)&(98.2, 93.9)  &(97.5, 93.0) \\  
\bottomrule
\end{tabular}%
\vspace{-1em}
\end{table}

\paragraph{Comparison with SOTA full-shot methods}
In this section, we are interested in the performance gap between AnomalyCLIP and the recently published SOTA full-shot methods, such as PatchCore~\citep{roth2022towards} and RD4AD~\citep{deng2022anomaly}. Since some datasets do not provide normal training data, we conduct experiments on six public datasets. AnomalyCLIP achieves comparable anomaly detection and segmentation performance compared to PatchCore and RD4AD, and it even outperforms them in some datasets. This illustrates that the generic prompt embeddings empower AnomalyCLIP to effectively capture the normality and abnormality so that AnomalyCLIP can surpass the performance boundary decided by the training data.

%%%%%%%%%%%%%%%%%%%%%%%%%%%%%%%%%%%%%%%%%%%%%%%%%%%%%%%%%%%%%%%%%%%%%%%%%%%%%%%%%%%%%%%%%%%%%%%%%%%%%

\begin{figure*}[]
  \centering
  \subfloat[]{\includegraphics[width=0.49\textwidth]{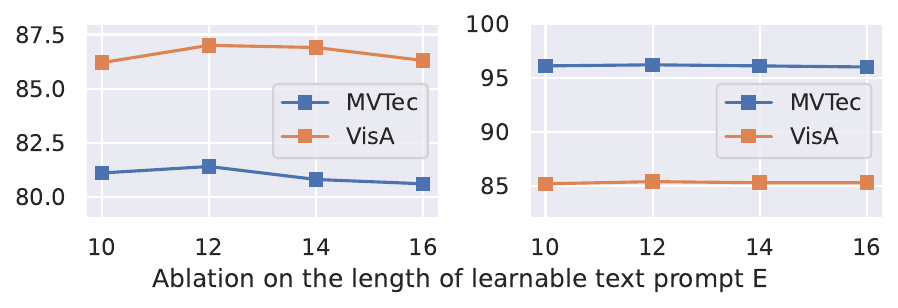}
  \label{length ablation}}
   \hfil
   \vspace{-0.6em}
  \subfloat[]{\includegraphics[width=0.49\textwidth]{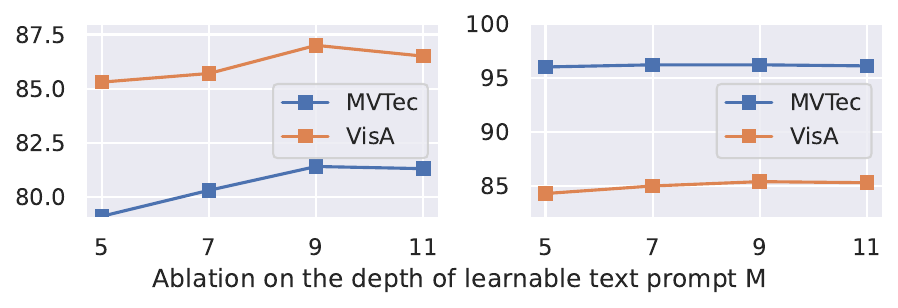}
  \label{depth ablation}}
     \hfil
   \vspace{-0.6em}
  \subfloat[]{\includegraphics[width=0.49\textwidth]{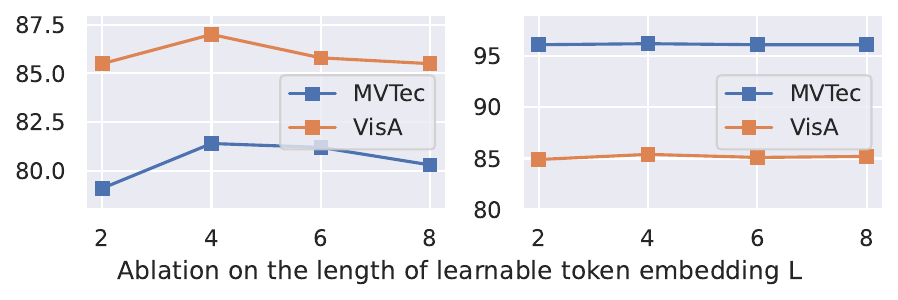}
  \label{length2 ablation}}
  %    \hfil
  \subfloat[]{\includegraphics[width=0.47\textwidth]{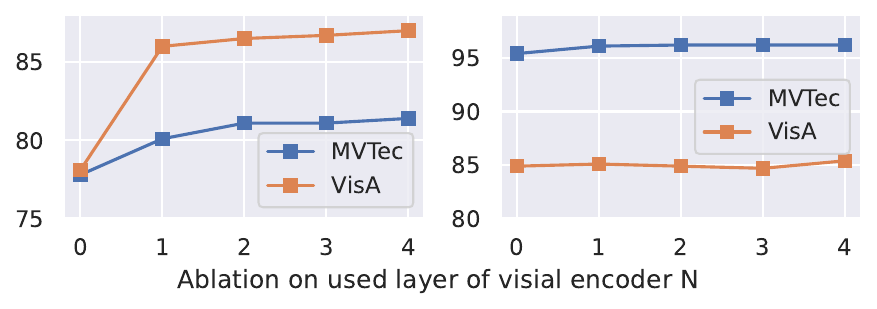}
  \label{layer ablation}}
\caption{Hyperparameter analysis. (a)$E$ ablation. (b) 
$M$ ablation. (c)$L$ ablation (d)$\lambda$ analysis. Pixel/image-level (AUPRO, AP) performances are shown on the left and right sides of each subplot, respectively.}
\label{Fig7: Hyparameters ablation}
\vspace{-1em}
\end{figure*}

\paragraph{Hyparameter analysis}
We study the length of learnable text prompts $E$, depth of learnable token embeddings $M$, length of learnable token embeddings $M$, and number of used layers in visual encoder $N$. As shown in Fig.~\ref{depth ablation}, we observe that the detection and segmentation performance initially improve with an increase in the value of $E$. However, within the range of lengths from 12 to 16, we notice a decline in performance, which suggests that excessively long learnable text prompts could involve redundant information. Therefore, an appropriate value for $E$, such as $E = 12$, is beneficial for accurate learning of object-agnostic text prompts. Besides, we also investigate the depth of the attached learnable token embeddings in Fig.~\ref{depth ablation}. The degree of refining of the initial text space becomes more pronounced as the depth increases, enabling more discriminative textual embeddings for normal and anomaly. However, the performance drops when the refinement is excessive and impairs the generalization of AnomalyCLIP, as seen in the case when $M$ equals 9. After selecting the depth, we proceed to investigate the influence of the length of learnable token embeddings. As illustrated in Fig.~\ref{length2 ablation}, we find that the length of token embeddings also involves a similar tradeoff between the model generalization and calibration of textual space. In Fig.~\ref{layer ablation}, AnomalyCLIP achieves the overall performance gain when $\lambda = 4$.

\paragraph{Prompt template ablation}
Here, we study the robustness of AnomalyCLIP to prior anomaly semantics in the object-agnostic text prompt template. We replace \verb'damaged' in the object-agnostic text prompt with other words having similar anomaly semantics, such as \verb'anomalous', \verb'flawed', \verb'defective', \verb'blemished'. The results are presented in Table~\ref{table: Ablation on the robustness of prior anomaly semantics on industrial domain.} and Table~\ref{table: Ablation on the robustness of prior anomaly semantics on medical domain.}. The steady results indicate that AnomalyCLIP is not sensitive to the prior anomaly semantics introduced by the object-agnostic text prompt template.

\begin{table}[]
\centering
\caption{Ablation on the robustness of the abnormality-related token in our prompt template on industrial defect datasets.}
\label{table: Ablation on the robustness of prior anomaly semantics on industrial domain.}
\tiny
\begin{tabular}{cccccccc}
\toprule
Task & Category &Datasets  & damaged & anomalous   & flawed & defective   & blemished  \\ \hline

\multirow{7}{*}{\makecell[c]{Image-level \\ (AUROC,  AP)}} & \makecell[c]{Obj \&texture}    &MVTec AD       & (91.5, 96.2)   & (91.4, 96.2)    & (91.3, 96.\textbf{}2)         & (91.4, 96.2)     & (91.5, 96.2)  \\
& \multirow{4}{*}{Obj}                                                                      & VisA          & (82.1, 85.4)   & (80.7, 84.5)    & (80.7, 84.5)         & (80.9, 84.6)     & (80.7, 84.5)   \\
&                                                                                           & MPDD          & (77.0, 82.0)   & (78.0, 83.9)    & (77.9, 83.6)         & (77.8, 83.5)     & (78.6. 84.1)     \\    
&                                                                                           & BTAD          & (88.3, 87.3)   & (84.8, 86.7)    & (85.2, 87.4)         & (84.8, 86.2)     & (85.9, 67.1)     \\     
&                                                                                           & SDD           & (84.7, 80.0)   & (82.3, 76.3)    & (82.6, 76.8)         & (82.8, 77.2)     & (82.7, 77.0)       \\
&  \multirow{2}{*}{Texture}                                                                 & DAGM          & (97.5, 92.3)   & (97.7, 92.6)    & (97.5, 92.4)         & (97.5, 92.3)     & (97.5, 92.4)       \\
&                                                                                           & DTD-Synthetic & (93.5, 97.0)   & (93.3, 96.9)    & (93.2, 96.9)         & (93.4, 97.0)     & (93.5, 97.0)      \\  \hline
                                                                                            
\multirow{7}{*}{\makecell[c]{Pixel-level \\ (AUROC, PRO)}} & \makecell[c]{Obj \&texture}    &MVTec AD       & (91.1, 81.4)   & (91.0, 81.4)    & (90.7, 81.4)         & (91.0, 81.7)     & (90.9, 81.2)  \\
& \multirow{4}{*}{Obj}                                                                      & VisA          & (95.5, 87.0)   & (95.5, 86.5)    & (95.5, 86.5)         & (95.5, 86.2)     & (95.6, 86.5)   \\
&                                                                                           & MPDD          & (96.5, 88.7)   & (96.6, 88.7)    & (96.7, 89.0)         & (96.7, 89.2)     & (96.6, 88.8)     \\    
&                                                                                           & BTAD          & (94.2, 74.8)   & (94.3, 74.3)    & (94.4, 75.1)         & (94.3, 75.2)     & (94.3, 73.7)     \\     
&                                                                                           & SDD           & (90.6, 67.8)   & (89.6, 66.8)    & (89.5, 66.5)         & (89.5, 64.8)     & (89.6, 64.6)       \\
&  \multirow{2}{*}{Texture}                                                                 & DAGM          & (95.6, 91.0)   & (95.6, 91.2)    & (95.6, 91.3)         & (95.5, 90.9)     & (95.6, 90.9)       \\
&                                                                                           & DTD-Synthetic & (97.9, 92.3)   & (97.9, 92.3)    & (97.9, 92.1)         & (97.9, 92.5)     & (97.9, 92.2)      \\  
\bottomrule
\end{tabular}%
\end{table}

\begin{figure*}[t]
    \centering
    \includegraphics[width=0.8\textwidth]{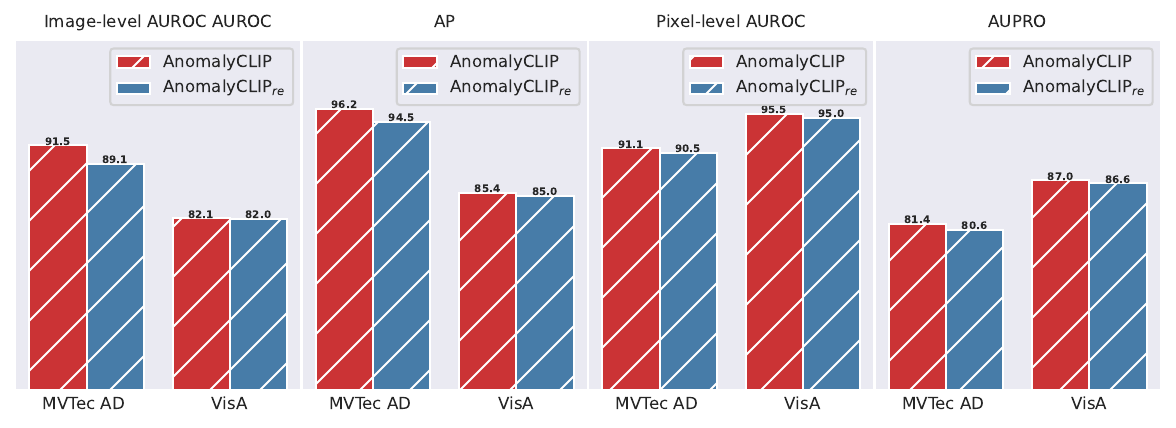}
    \caption{Object ablation.}
    \label{Object ablation.}
\end{figure*}
\paragraph{Object ablation}
To investigate what the object-agnostic text prompts have learned, we replace \verb'object' in object-agnostic text prompts with specific target \verb'[cls]', resulting in AnomalyCLIP$_{re}$. In Fig.~\ref{Object ablation.}, AnomalyCLIP$_{re}$ still performs well in ZSAD, even as we block out the object semantics during fine-tuning. This suggests that the knowledge learned by object-agnostic text prompts is the underlying anomaly patterns, allowing them to provide discriminative textual embeddings even when specific object semantics are incorporated. Furthermore, compared to AnomalyCLIP, AnomalyCLIP$_{re}$ shows a performance decay, which can be attributed to the inclusion of redundant/noisy object semantics. These results once again demonstrate the generalization ability of object-agnostic prompt learning.

\begin{table}[]
\centering
\caption{Ablation on the robustness of the abnormality-related token in our prompt template on medical image datasets.}

\label{table: Ablation on the robustness of prior anomaly semantics on medical domain.}
\tiny
\begin{tabular}{cccccccc}
\toprule
Task & Category &Datasets  & damaged & anomalous   & flawed & defective   & blemished  \\ \hline

\multirow{4}{*}{\makecell[c]{Image-level \\  (AUROC, AP)}} & \multirow{3}{*}{Brain}         & HeadCT        & (93.4, 91.6)    & (93.1, 90.6)    & (93.3, 90.8)         & (93.5, 91.0)     & (93.8, 91.5)        \\ 
&                                                                                           & BrainMRI      & (90.3, 92.2)    & (87.8, 90.4)    & (87.7, 90.0)         & (88.3, 90.5)     & (88.6, 90.7)        \\  
&                                                                                           & Br35H         & (94.6, 94.7)    & (93.1, 93.0)    & (92.9, 92.8)         & (93.1, 93.0)     & (93.2, 93.1)       \\   
& Chest                                                                                     & COVID-19      & (80.1, 58.7)    & (80.0, 58.5)    & (80.2, 58.8)         & (80.6, 59.0)     & (82.1, 61.4)     \\  \hline   
\multirow{6}{*}{\makecell[c]{Pixel-level \\ (AUROC, PRO)}} &  Skin                          & ISIC          & (89.7, 78.4)    & (90.1, 80.1)    & (90.1, 80.1)         & (90.4, 81.0)     & (90.2, 80.6)        \\
& \multirow{4 }{*}{Colon}                                                                   & CVC-ColonDB   & (81.9, 71.3)    & (82.2, 71.5)    & (82.3, 71.6)         & (82.1, 71.1)     & (82.2, 71.5)        \\
&                                                                                           & CVC-ClinicDB  & (82.9, 67.8)    & (83.0, 68.1)    & (83.1, 68.4)         & (82.9, 67.9)     & (83.1, 68.2)        \\
&                                                                                           & Kvasir        & (78.9, 45.6)    & (79.4, 45.1)    & (79.4, 45.2)         & (79.3, 44.9)     & (79.5, 45.8)       \\   
&                                                                                           & Endo          & (84.1, 63.6)    & (84.3, 63.5)    & (84.2, 63.5)         & (84.2, 62.9)     & (84.3, 63.4)     \\     
& Thyroid                                                                                   & TN3K          & (81.5, 50.4)    & (81.5, 51.7)    & (81.3, 50.9)         & (81.3, 50.3)     & (81.6, 51.1)     \\

\bottomrule
\end{tabular}%
\vspace{-2em}
\end{table}

\section{Visualization}
\label{Visualization}
\begin{figure*}[h]
    \centering
  \includegraphics[width=1\textwidth]{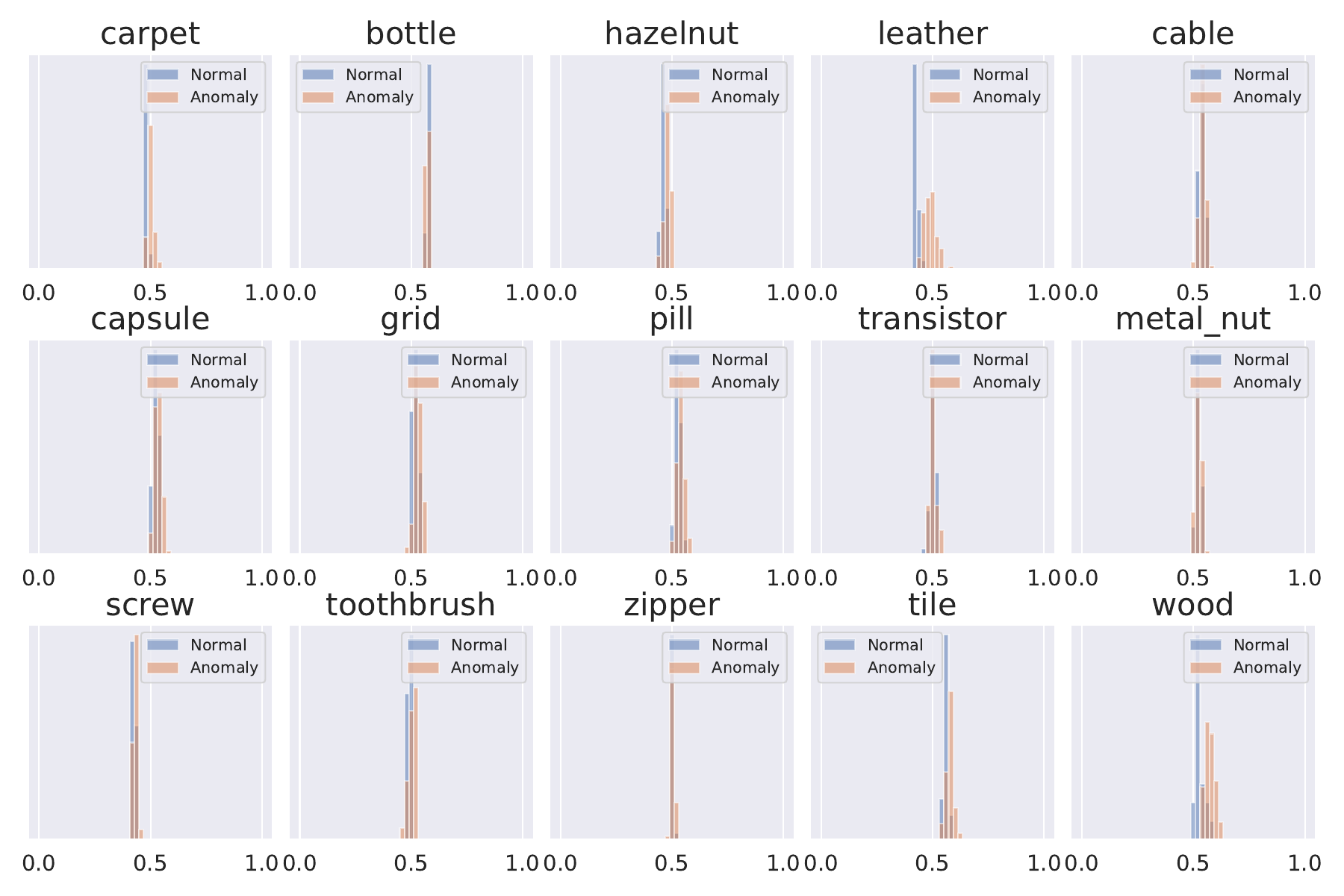}
    \caption{Similarity scores of CLIP on MVTec AD. Each sub-figure represents the visualization of one object.}
    %\\vspace{-1em}
    \label{mvtec_class_level_similarity_map_all}
\end{figure*}

\begin{figure*}[h]
    \centering
  \includegraphics[width=1\textwidth]{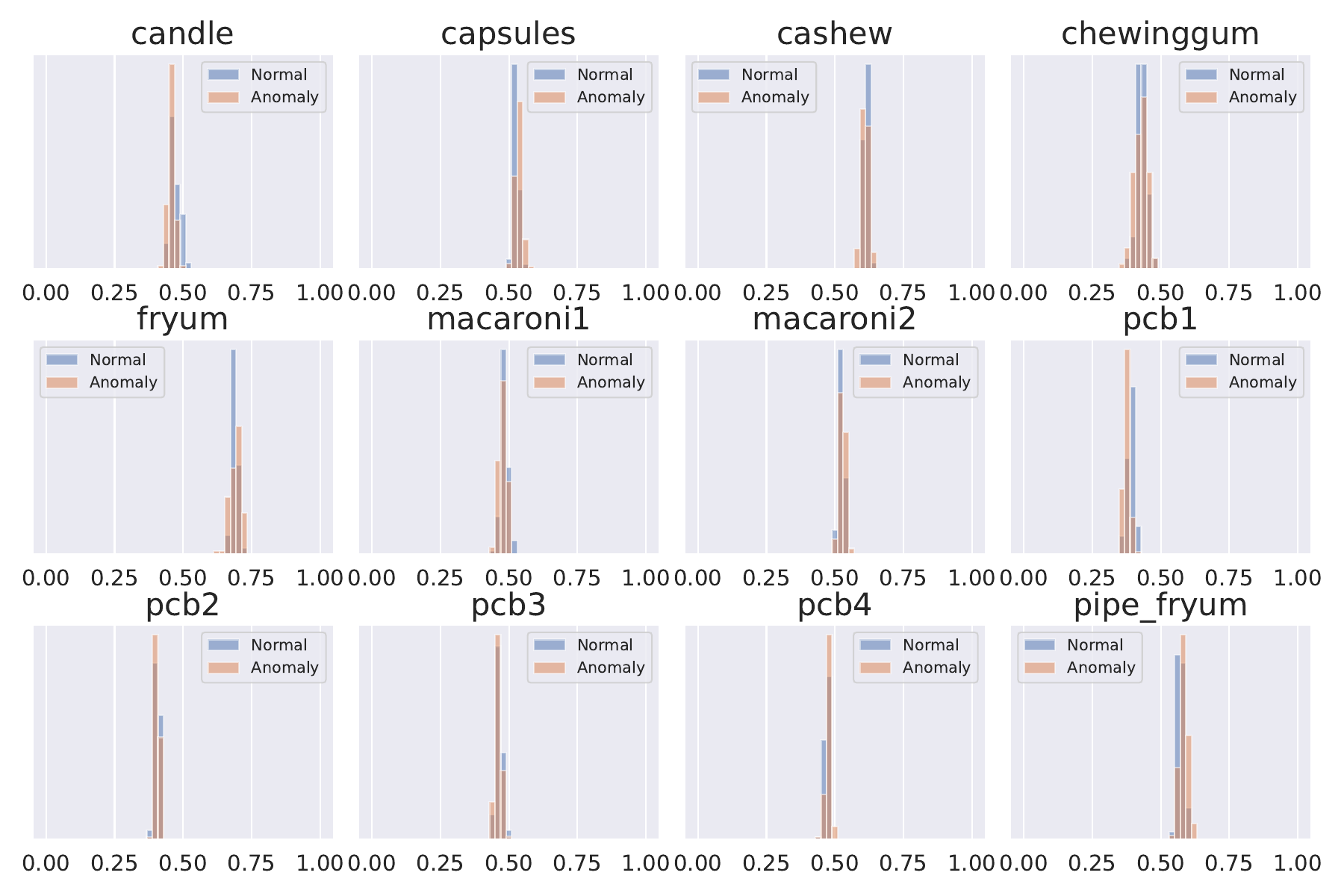}
    \caption{Similarity scores of CLIP on VisA. Each sub-figure represents the visualization of one object.}
    %\\vspace{-1em}
    \label{visa_class_level_similarity_map_all}
\end{figure*}

\begin{figure*}[h]
    \centering
  \includegraphics[width=1\textwidth]{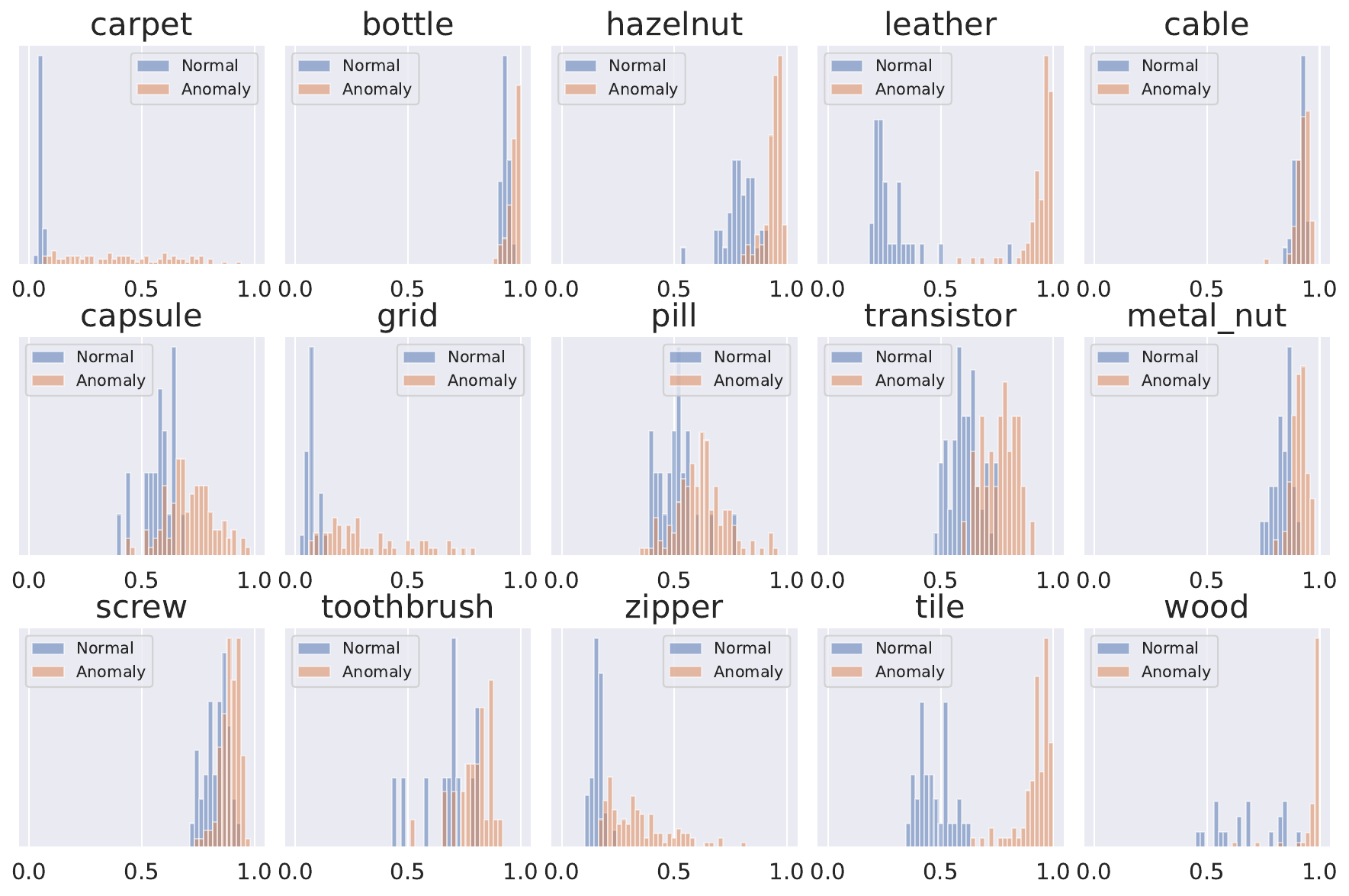}
    \caption{Similarity scores of AnomalyCLIP on MVTec AD. Each sub-figure represents the visualization of one object.}
    %\\vspace{-1em}
    \label{AnomalyCLIP_mvtec_class_level_similarity_map_all}
\end{figure*}

\begin{figure*}[h]
    \centering
  \includegraphics[width=1\textwidth]{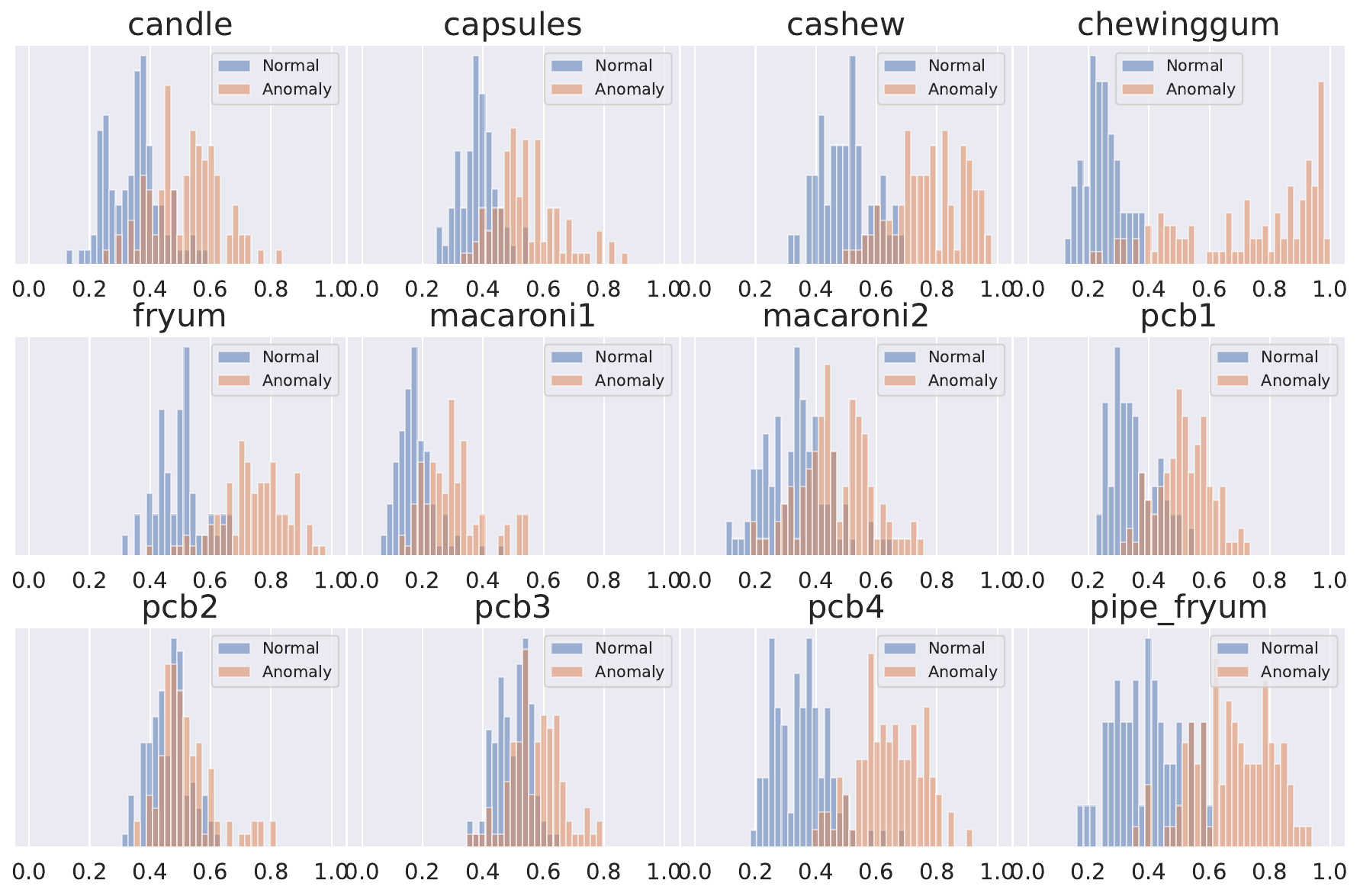}
    \caption{Similarity scores of AnomalyCLIP on VisA. Each sub-figure represents the visualization of one object.}
    %\\vspace{-1em}
    \label{AnomalyCLIP_visa_class_level_similarity_map_all}
\end{figure*}

\paragraph{Similarity score between textual and visual embeddings.}
We present visualizations of the similarity scores generated by both CLIP and AnomalyCLIP. These visualizations aim to provide an intuitive illustration of the effective adaptation made by AnomalyCLIP in comparison to CLIP. As shown in Fig.~\ref{mvtec_class_level_similarity_map_all} and Fig.~\ref{visa_class_level_similarity_map_all}, we present the similarity score of CLIP on MVTec AD and VisA. The normal and anomaly scores are severely overlapped. Further, the range of scores is centered at 0.5. These show that the textual and visual space of CLIP originally aligned for object semantics are not desired for ZSAD. Also, we visualize the similarity scores 
 of AnomalyCLIP in Fig.~\ref{AnomalyCLIP_mvtec_class_level_similarity_map_all} and Fig.~\ref{AnomalyCLIP_visa_class_level_similarity_map_all}. Compared to CLIP, there is a significant overlap between the scores assigned to normal and anomaly instances, and at the same time, the score range is considerably wider. These results indicate that AnomalyCLIP achieves a significant improvement in adapting CLIP to ZSAD.

\paragraph{Anomaly score map for different datasets.}
In addition to the similarity score for anomaly classification, we also visualize the anomaly score maps to present the strong anomaly segmentation ability of AnomalyCLIP. Specifically, we visualize the industrial object class: hazelnut, pill, and screw from MVTec AD; candle, chewinggum, capsule, cashew, pcb, and pip fryum from Visa; bracket, metal plate, and tube from MPDD. We also visualize the industrial texture: grid, leather, carpet, tile, wood, and zipper. In addition, we visualize the segmentation in medical domain across photography, endoscopy, and radiology images: skin cancer detection from ISIC; thyroid nodule detection from TN3K; colon polyp detection from Kvasir; brain tumor detection from Br35H.   

\begin{figure*}[t]
    \centering
  \includegraphics[width=1\textwidth]{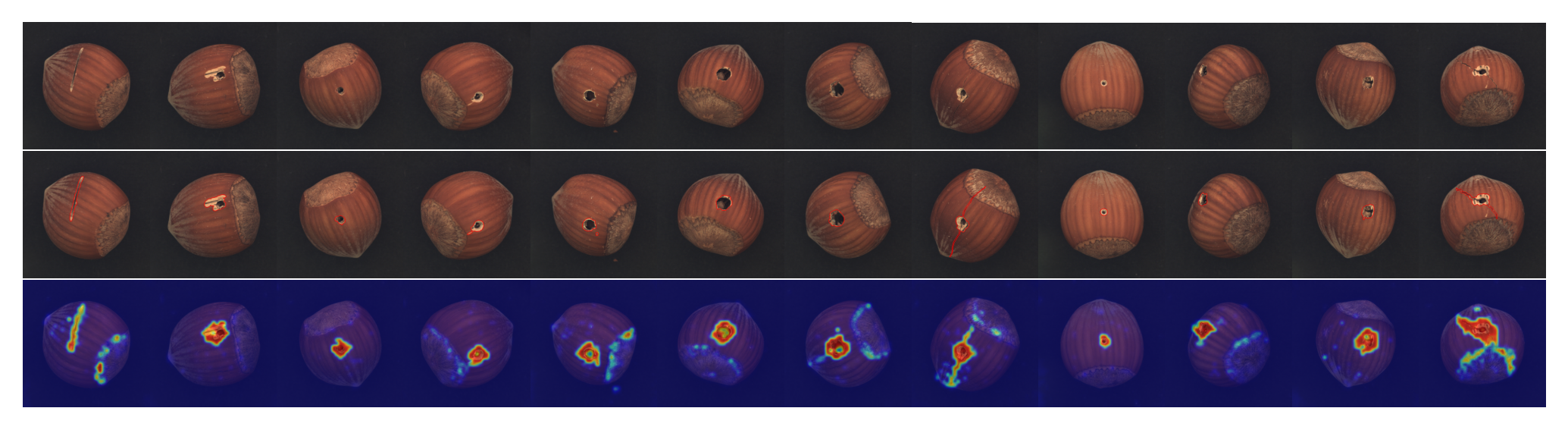}
    \caption{Anomaly score maps for the data subset, hazelnut, in MVTec AD. The first row represents the input, and we circle the anomaly regions in the second row. The last row presents the segmentation results from AnomalyCLIP.}
    \vspace{-1em}
    \label{}
\end{figure*}

\begin{figure*}[t]
    \centering
  \includegraphics[width=1\textwidth]{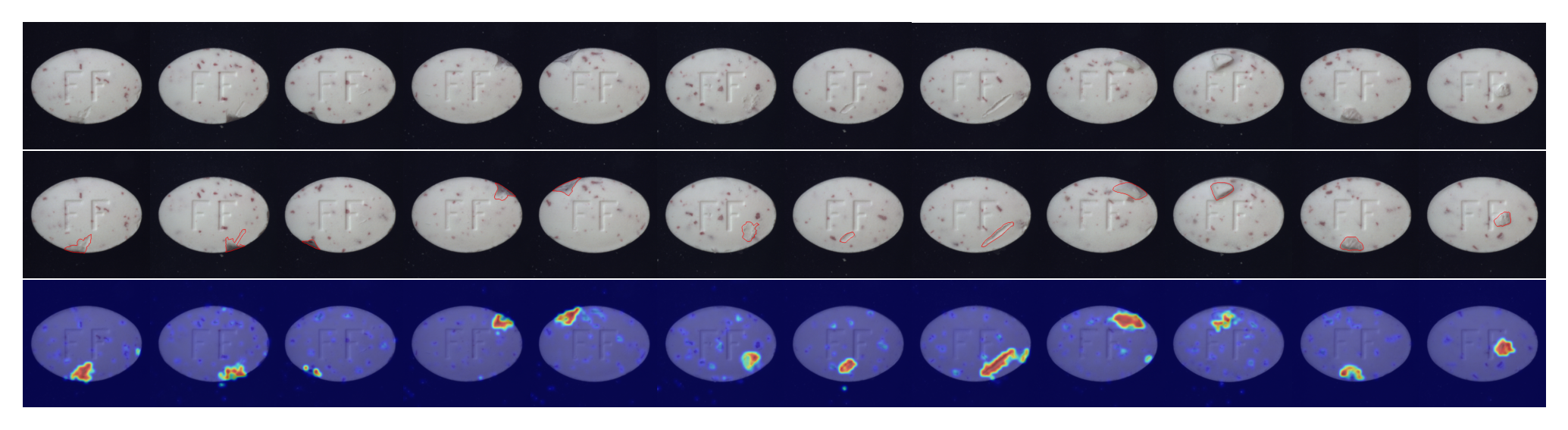}
    \caption{Anomaly score maps for the data subset, pill, in MVTec AD. The first row represents the input, and we circle the anomaly regions in the second row. The last row presents the segmentation results from AnomalyCLIP.}
    \vspace{-1em}
    \label{}
\end{figure*}

\begin{figure*}[t]
    \centering
  \includegraphics[width=1\textwidth]{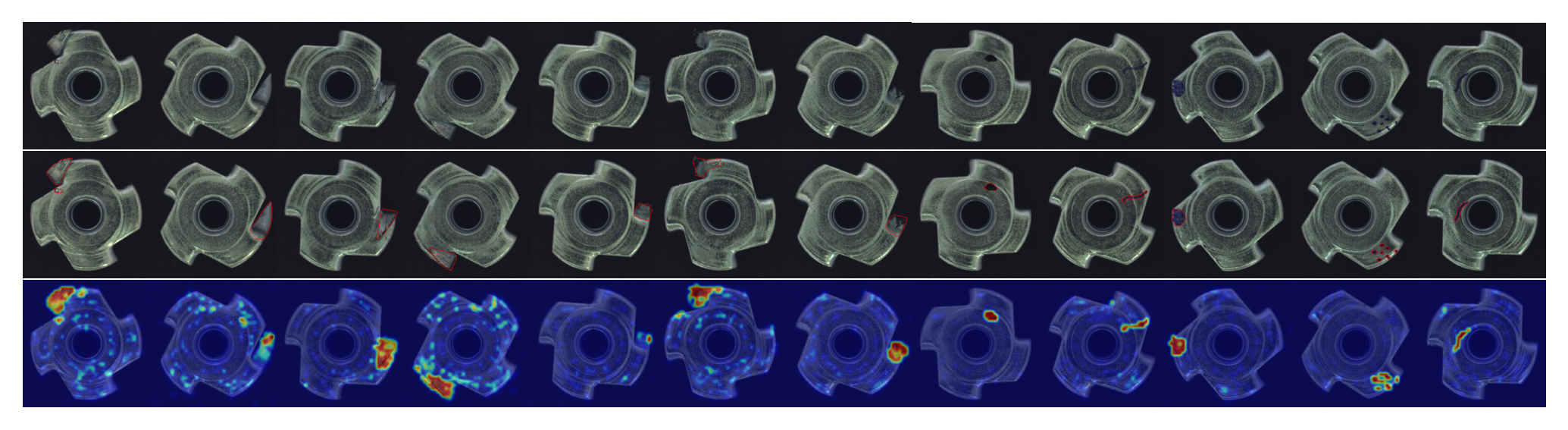}
    \caption{Anomaly score maps for the data subset, metal nut, in MVTec AD. The first row represents the input, and we circle the anomaly regions in the second row. The last row presents the segmentation results from AnomalyCLIP.}
    \vspace{-1em}
    \label{}
\end{figure*}

\begin{figure*}[t]
    \centering
  \includegraphics[width=1\textwidth]{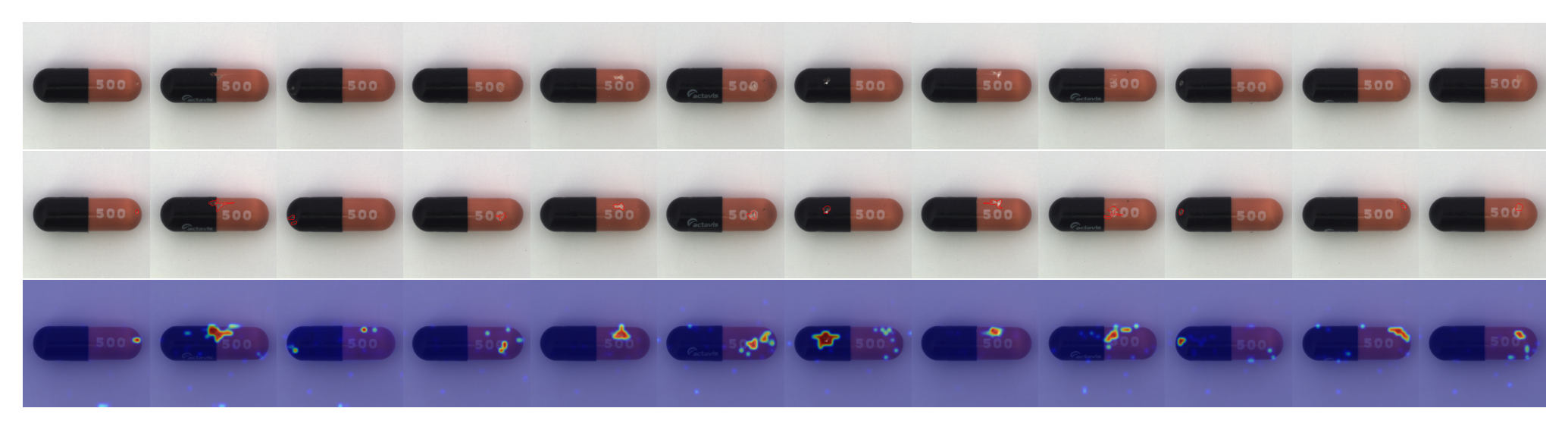}
    \caption{Anomaly score maps for the data subset, capsule, in MVTec AD. The first row represents the input, and we circle the anomaly regions in the second row. The last row presents the segmentation results from AnomalyCLIP.}
    \vspace{-1em}
    \label{}
\end{figure*}

\begin{figure*}[t]
    \centering
  \includegraphics[width=1\textwidth]{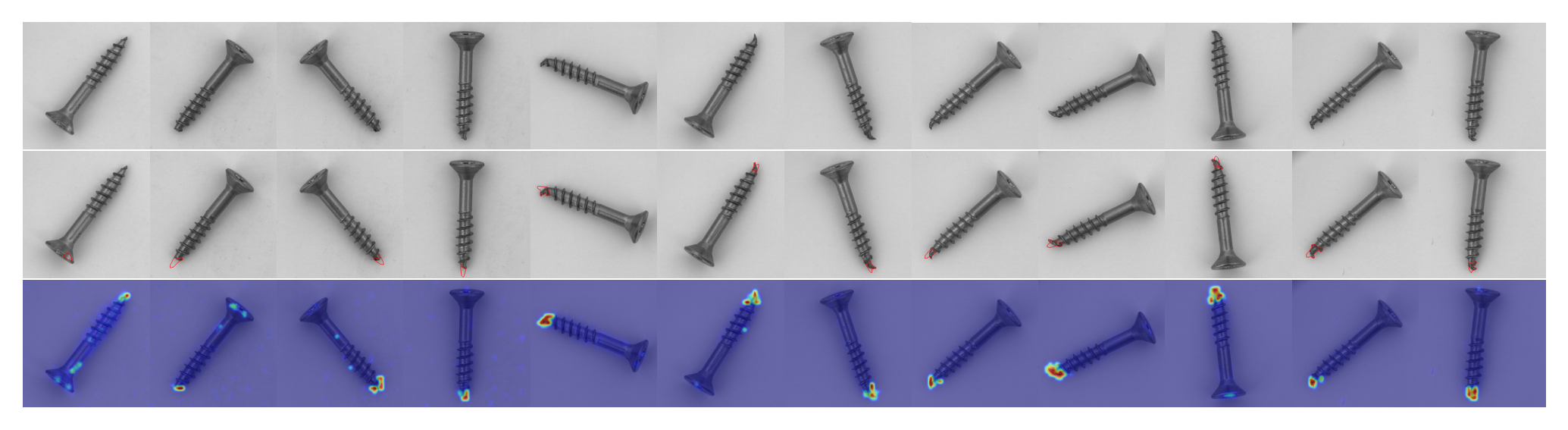}
    \caption{Anomaly score maps for the data subset, screw, in MVTec AD. The first row represents the input, and we circle the anomaly regions in the second row. The last row presents the segmentation results from AnomalyCLIP.}
    \vspace{-1em}
    \label{}
\end{figure*}

\begin{figure*}[t]
    \centering
  \includegraphics[width=1\textwidth]{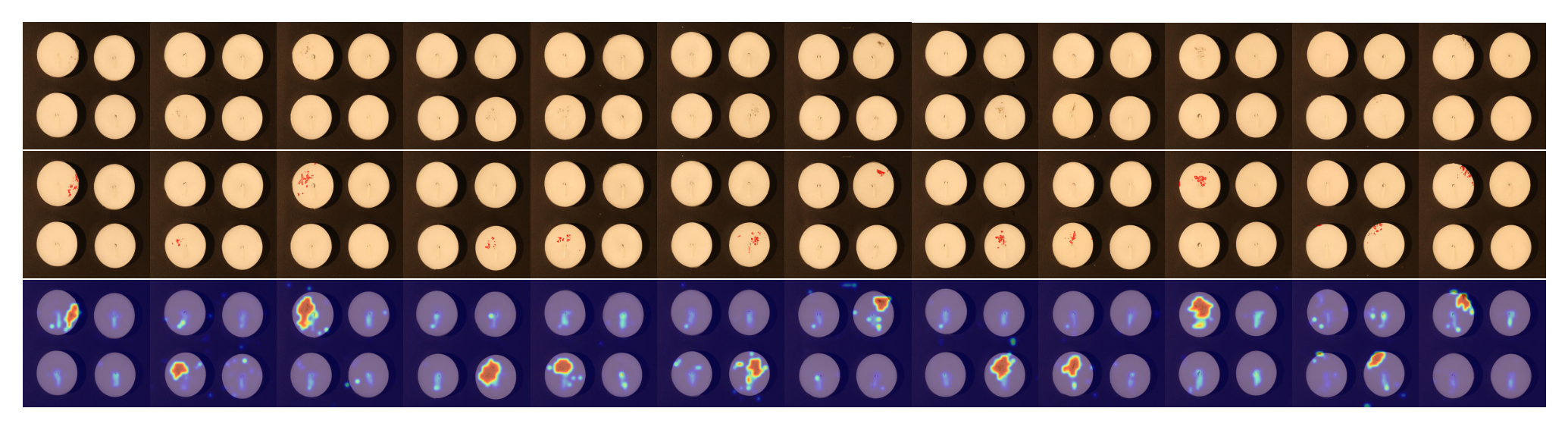}
    \caption{Anomaly score maps for the data subset candle. The first row represents the input, and we circle the anomaly regions in the second row. The last row presents the segmentation results from AnomalyCLIP.}
    \vspace{-1em}
    \label{}
\end{figure*}

\begin{figure*}[t]
    \centering
  \includegraphics[width=1\textwidth]{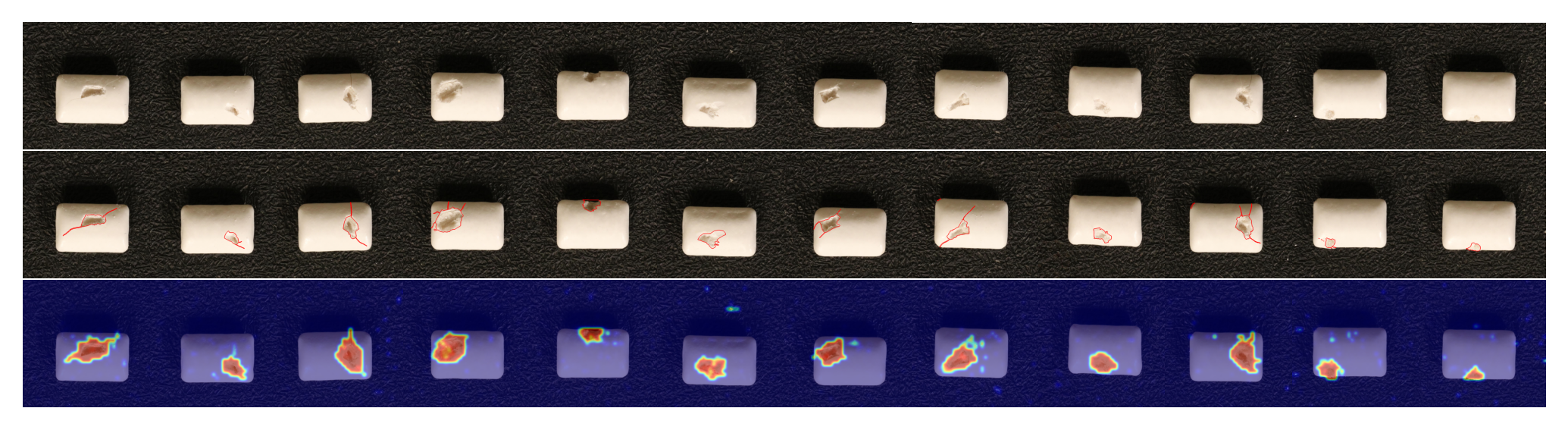}
    \caption{Anomaly score maps for the data subset chewinggum. The first row represents the input, and we circle the anomaly regions in the second row. The last row presents the segmentation results from AnomalyCLIP.}
    \vspace{-1em}
    \label{}
\end{figure*}

\begin{figure*}[t]
    \centering
  \includegraphics[width=1\textwidth]{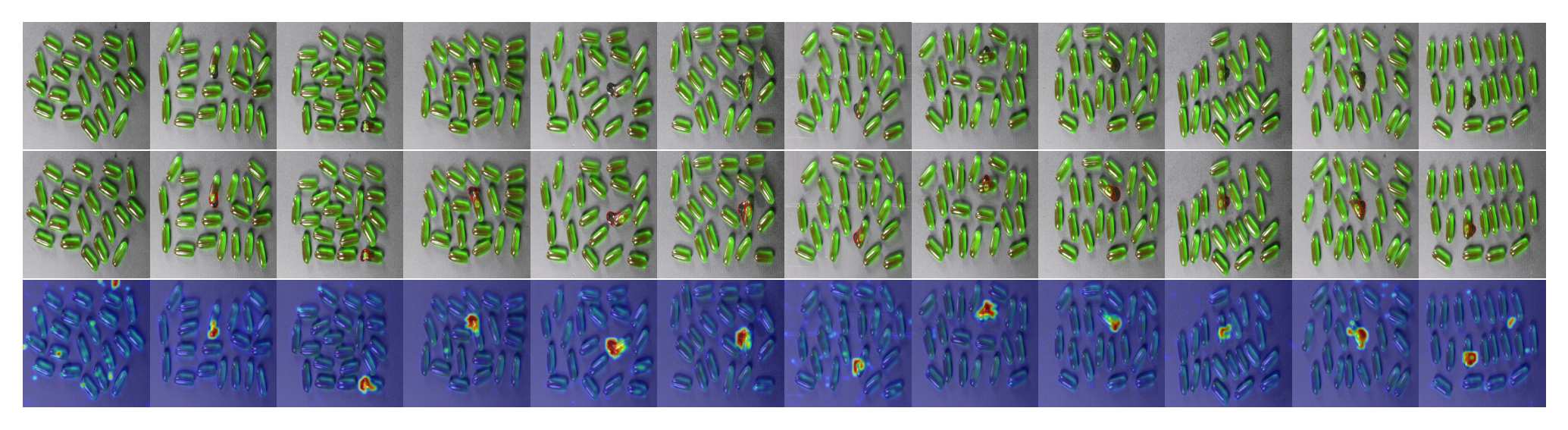}
    \caption{Anomaly score maps for the data subset capusle. The first row represents the input, and we circle the anomaly regions in the second row. The last row presents the segmentation results from AnomalyCLIP.}
    \vspace{-1em}
    \label{}
\end{figure*}

\begin{figure*}[t]
    \centering
  \includegraphics[width=1\textwidth]{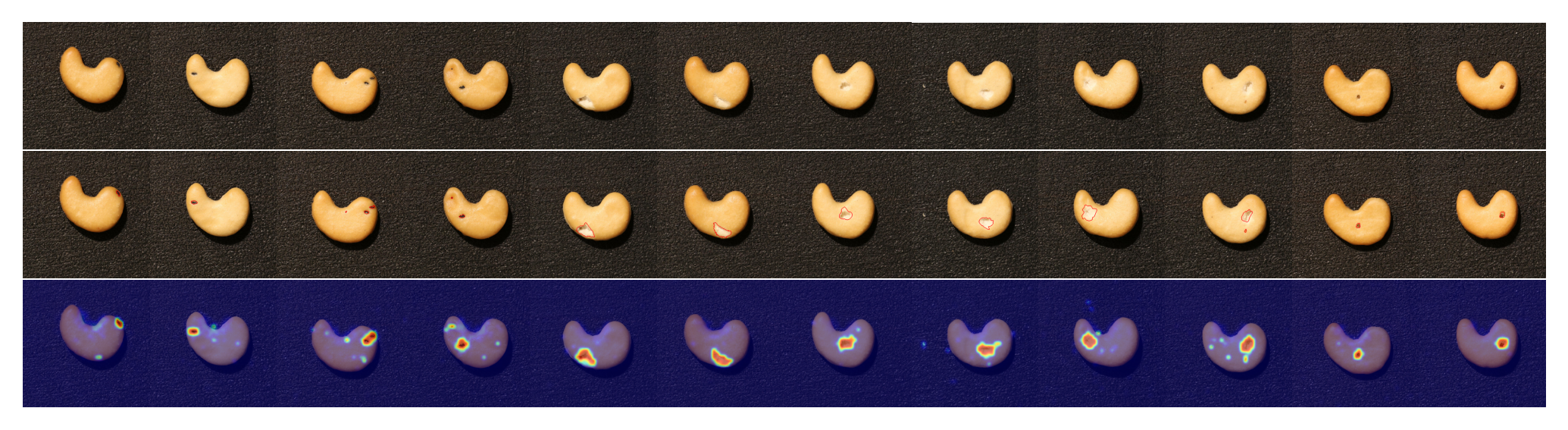}
    \caption{Anomaly score maps for the data subset cashew. The first row represents the input, and we circle the anomaly regions in the second row. The last row presents the segmentation results from AnomalyCLIP.}
    \vspace{-1em}
    \label{}
\end{figure*}

\begin{figure*}[t]
    \centering
  \includegraphics[width=1\textwidth]{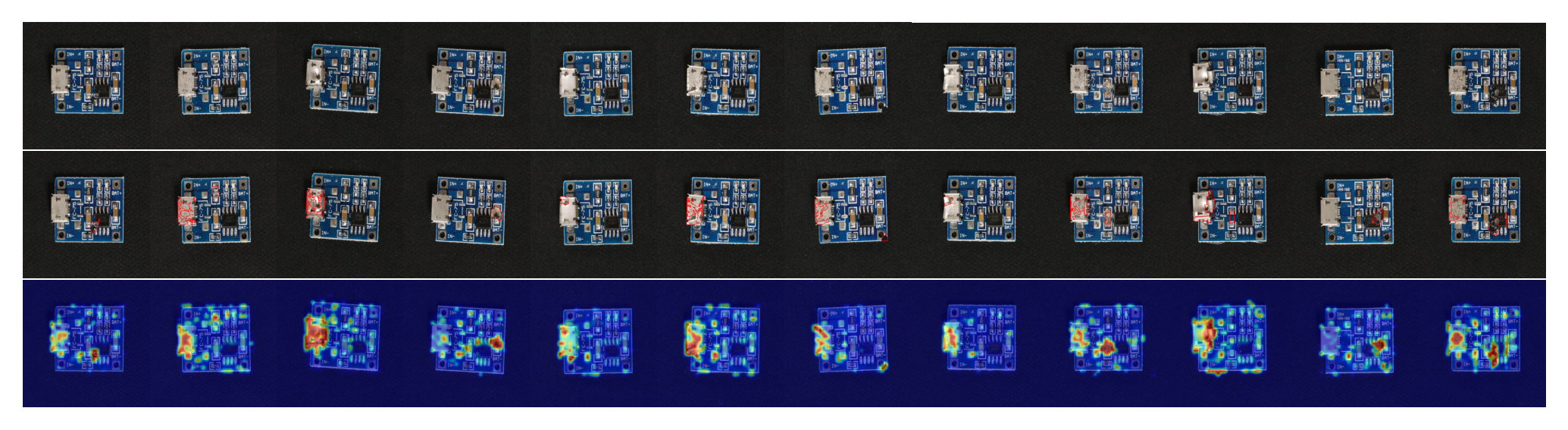}
    \caption{Anomaly score maps for the data subset pcb. The first row represents the input, and we circle the anomaly regions in the second row. The last row presents the segmentation results from AnomalyCLIP.}
    \vspace{-1em}
    \label{}
\end{figure*}

\begin{figure*}[t]
    \centering
  \includegraphics[width=1\textwidth]{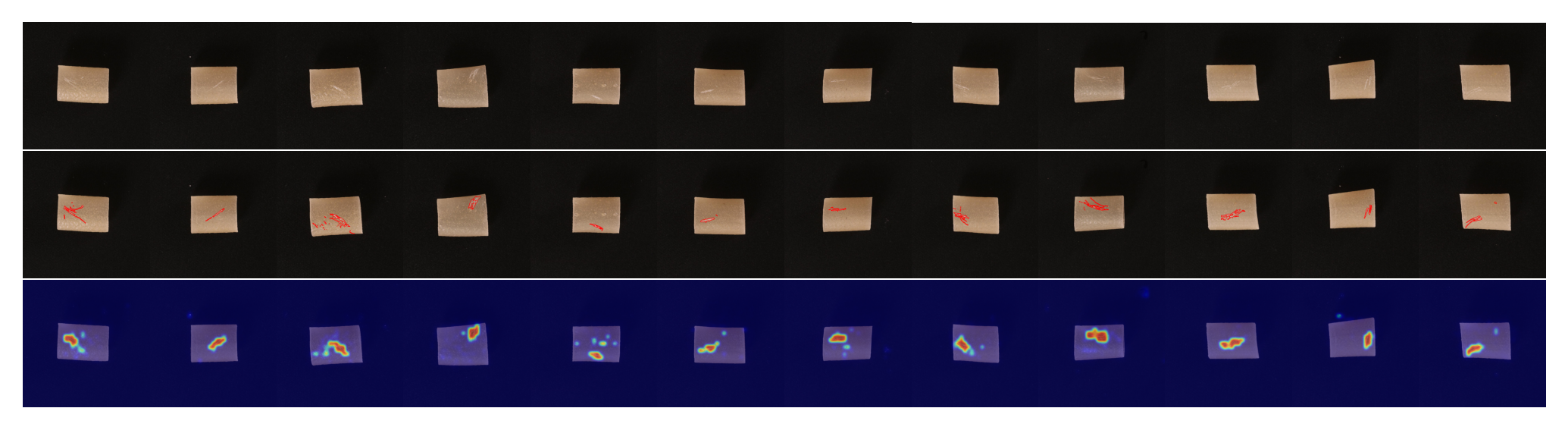}
    \caption{Anomaly score maps for the data subset pip fryum. The first row represents the input, and we circle the anomaly regions in the second row. The last row presents the segmentation results from AnomalyCLIP.}
    \vspace{-1em}
    \label{}
\end{figure*}

\begin{figure*}[t]
    \centering
  \includegraphics[width=1\textwidth]{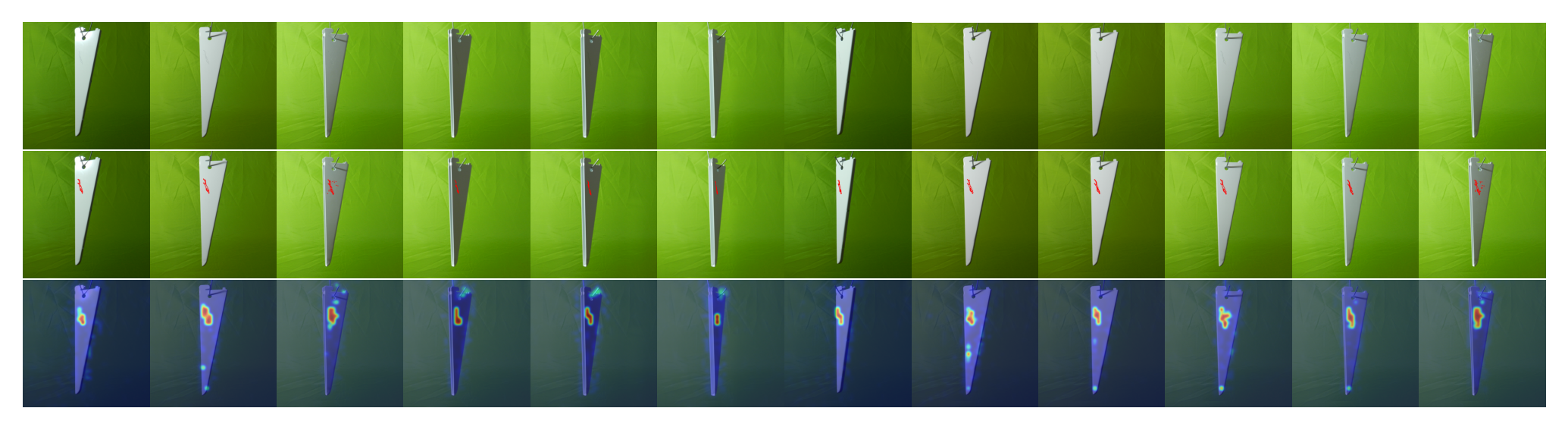}
    \caption{Similarity scores for the data subset bracket. The first row represents the input, and we circle the anomaly regions in the second row. The last row presents the segmentation results from AnomalyCLIP.}
    \vspace{-1em}
    \label{}
\end{figure*}

\begin{figure*}[t]
    \centering
  \includegraphics[width=1\textwidth]{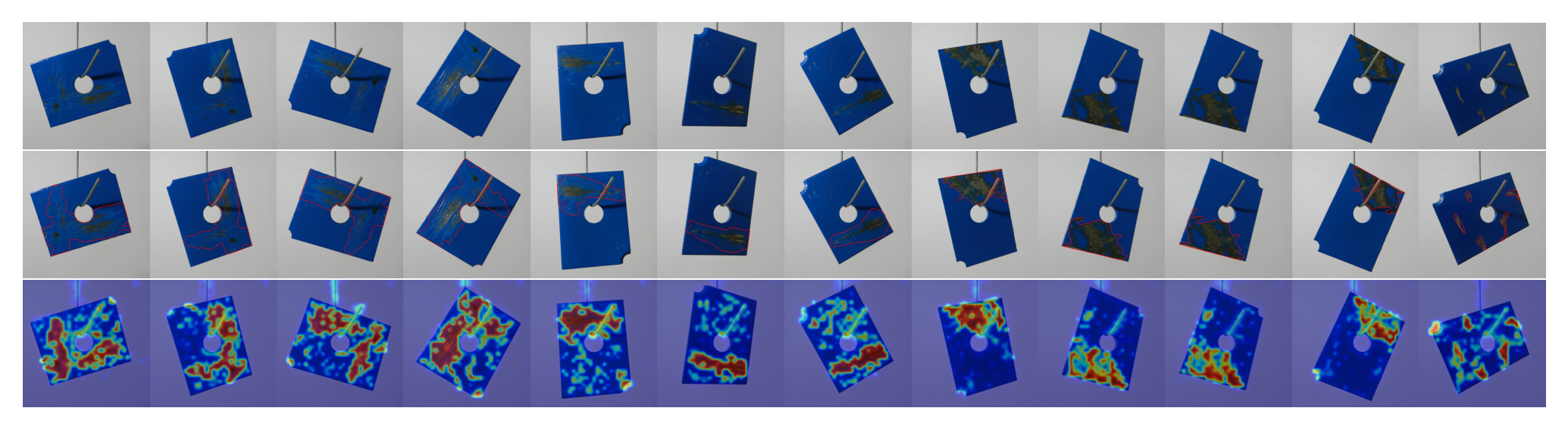}
    \caption{Anomaly score maps for the data subset metal plate. The first row represents the input, and we circle the anomaly regions in the second row. The last row presents the segmentation results from AnomalyCLIP.}
    \vspace{-1em}
    \label{}
\end{figure*}

\begin{figure*}[t]
    \centering
  \includegraphics[width=1\textwidth]{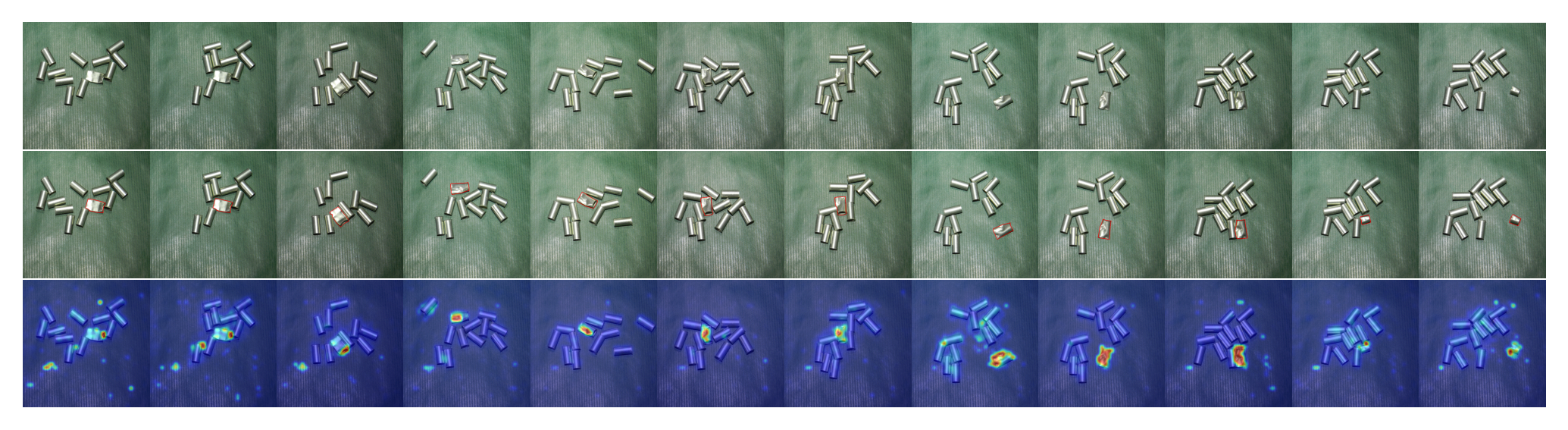}
    \caption{Anomaly score maps for the data subset tube. The first row represents the input, and we circle the anomaly regions in the second row. The last row presents the segmentation results from AnomalyCLIP.}
    \vspace{-1em}
    \label{}
\end{figure*}

\begin{figure*}[t]
    \centering
  \includegraphics[width=1\textwidth]{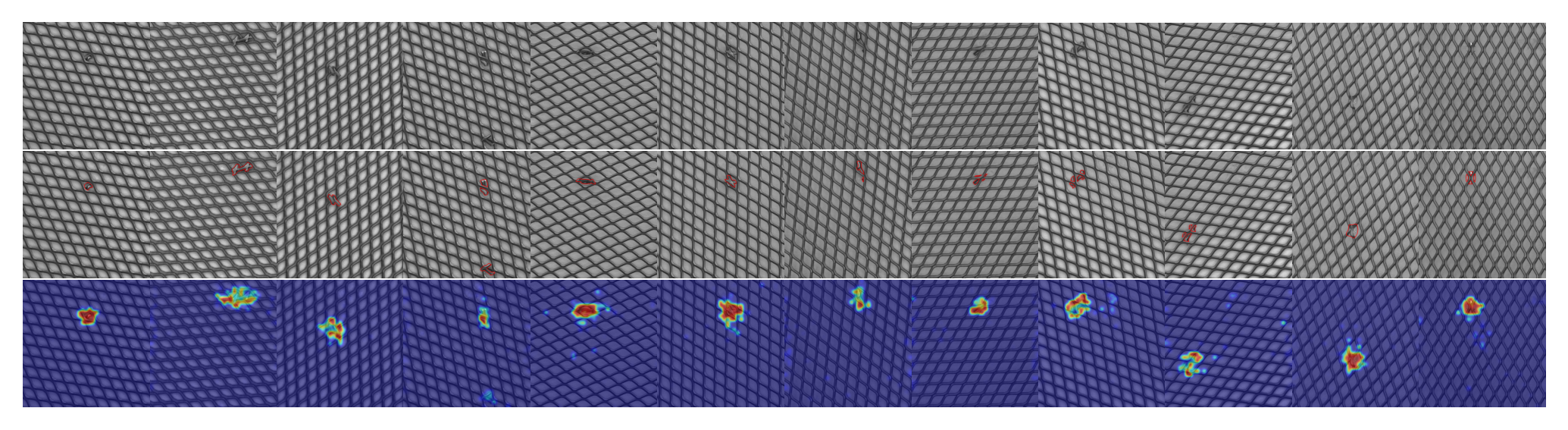}
    \caption{Anomaly score maps for the data subset grid. The first row represents the input, and we circle the anomaly regions in the second row. The last row presents the segmentation results from AnomalyCLIP.}
    \vspace{-1em}
    \label{}
\end{figure*}

\begin{figure*}[t]
    \centering
  \includegraphics[width=1\textwidth]{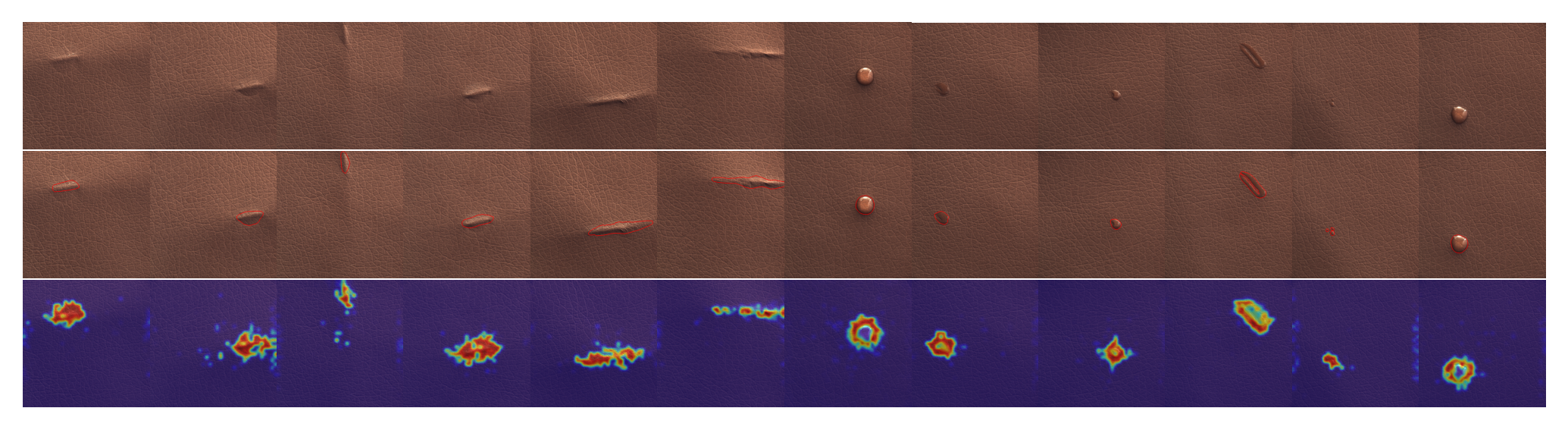}
    \caption{Anomaly score maps for the data subset leather. The first row represents the input, and we circle the anomaly regions in the second row. The last row presents the segmentation results from AnomalyCLIP.}
    \vspace{-1em}
    \label{}
\end{figure*}

\begin{figure*}[t]
    \centering
  \includegraphics[width=1\textwidth]{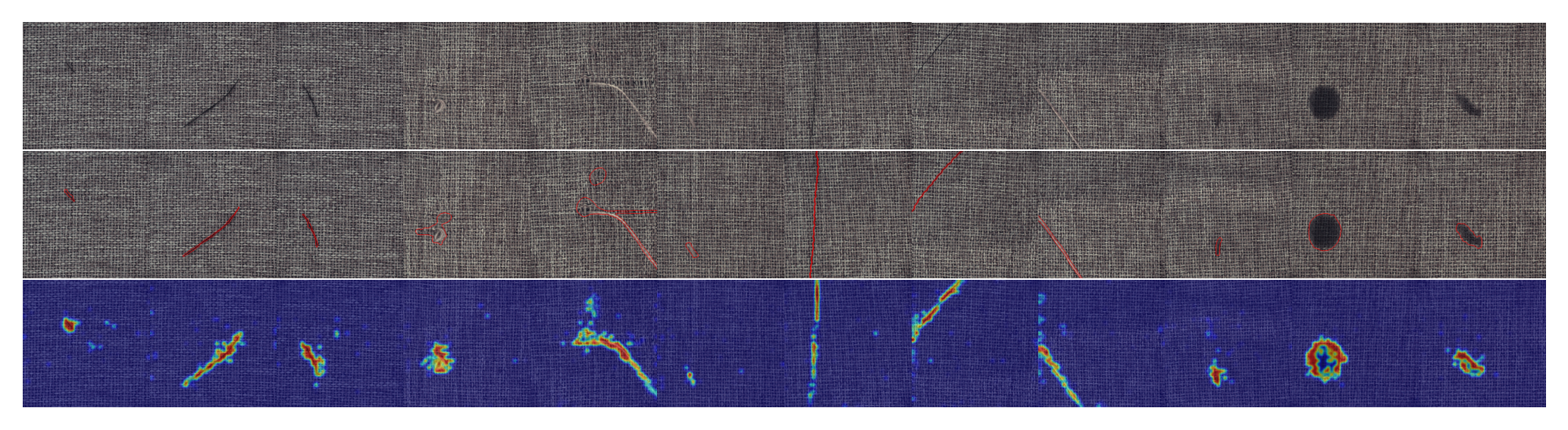}
    \caption{Anomaly score maps for the data subset carpet. The first row represents the input, and we circle the anomaly regions in the second row. The last row presents the segmentation results from AnomalyCLIP.}
    \vspace{-1em}
    \label{}
\end{figure*}

\begin{figure*}[t]
    \centering
  \includegraphics[width=1\textwidth]{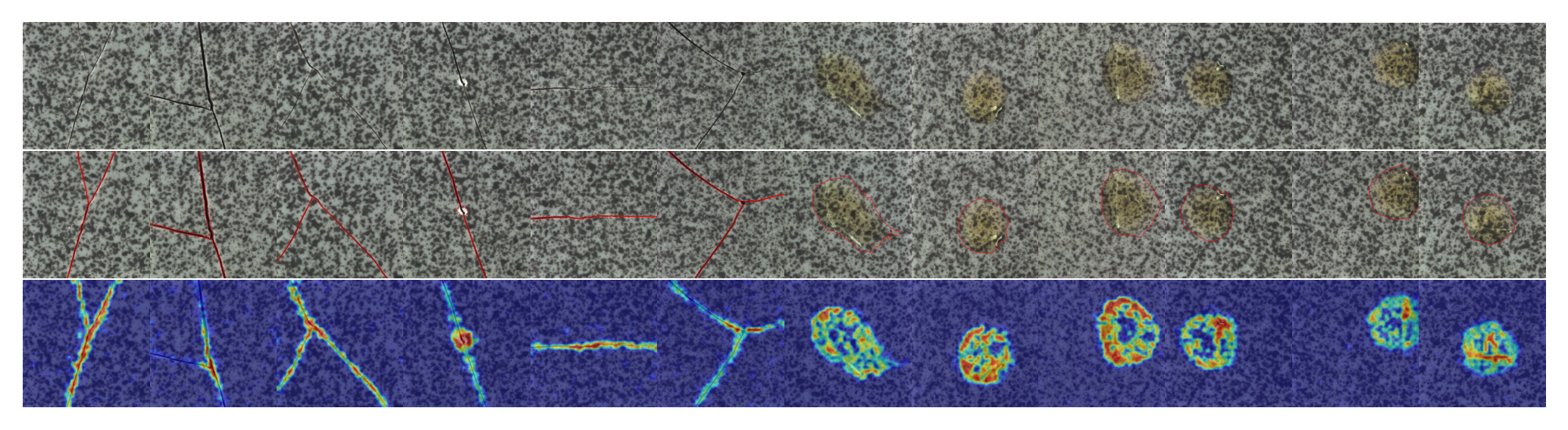}
    \caption{Anomaly score maps for the data subset tile. The first row represents the input, and we circle the anomaly regions in the second row. The last row presents the segmentation results from AnomalyCLIP.}
    \vspace{-1em}
    \label{}
\end{figure*}

\begin{figure*}[t]
    \centering
  \includegraphics[width=1\textwidth]{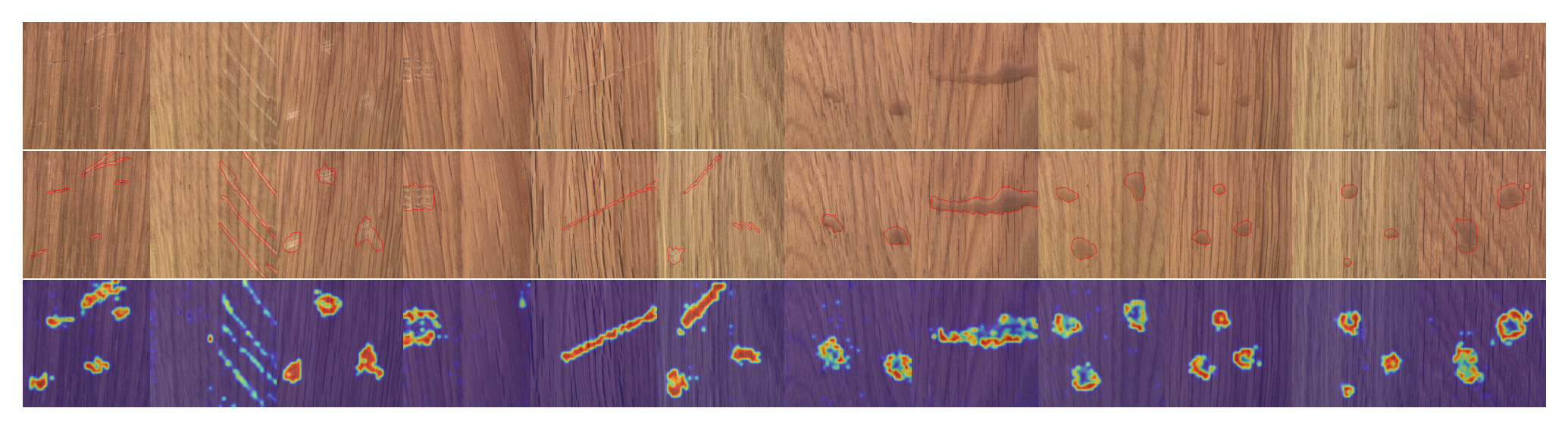}
    \caption{Anomaly score maps for the data subset wood. The first row represents the input, and we circle the anomaly regions in the second row. The last row presents the segmentation results from AnomalyCLIP.}
    \vspace{-1em}
    \label{}
\end{figure*}

\begin{figure*}[t]
    \centering
  \includegraphics[width=1\textwidth]{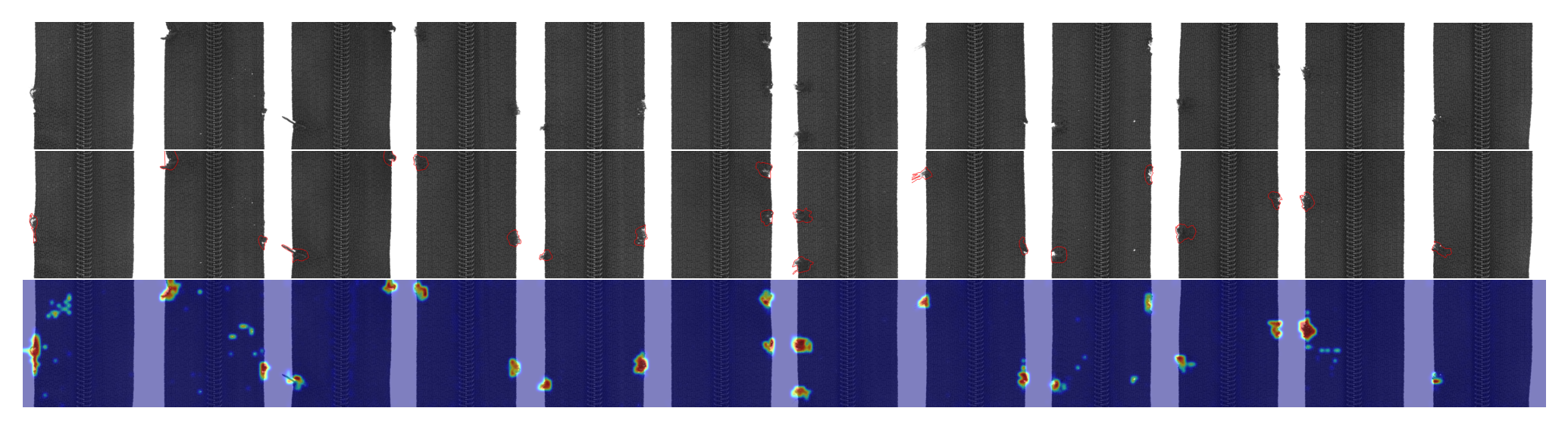}
    \caption{Anomaly score maps for the data subset zipper. The first row represents the input, and we circle the anomaly regions in the second row. The last row presents the segmentation results from AnomalyCLIP.}
    \vspace{-1em}
    \label{}
\end{figure*}

\begin{figure*}[t]
    \centering
  \includegraphics[width=1\textwidth]{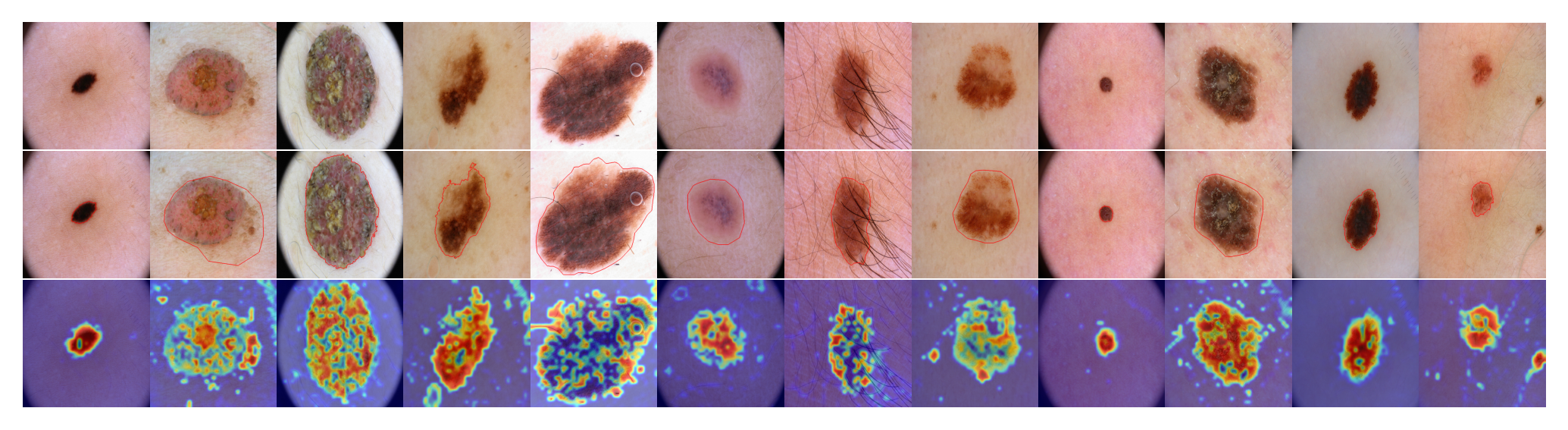}
    \caption{Similarity scores for the data subset skin.}
    \vspace{-1em}
    \label{}
\end{figure*}

\begin{figure*}[t]
    \centering
  \includegraphics[width=1\textwidth]{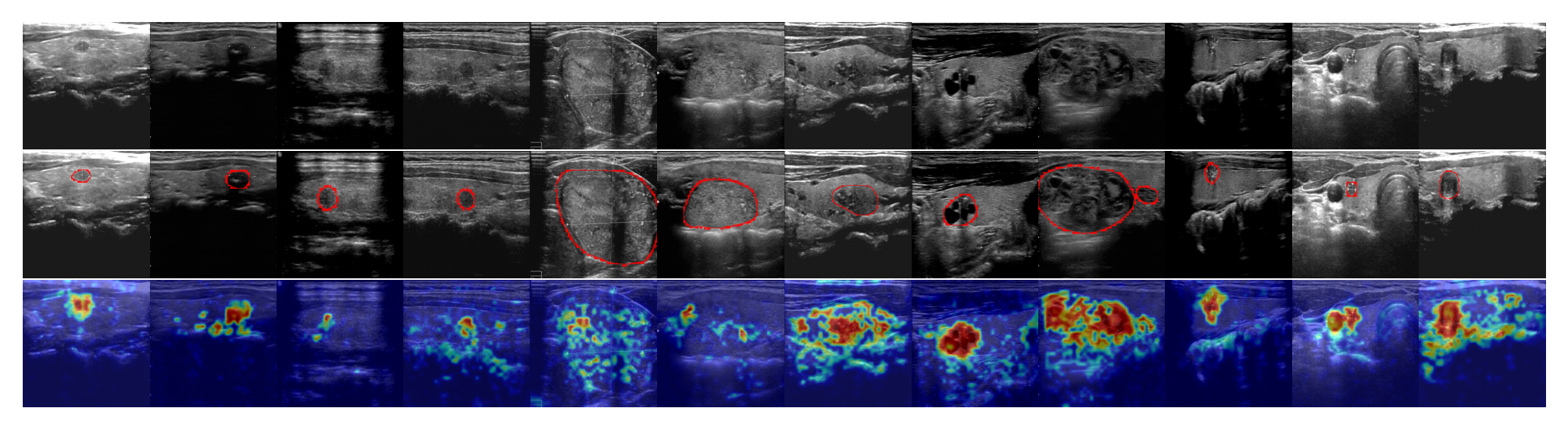}
    \caption{Anomaly score maps for the data subset thyroid. The first row represents the input, and we circle the anomaly regions in the second row. The last row presents the segmentation results from AnomalyCLIP.}
    \vspace{-1em}
    \label{}
\end{figure*}

\begin{figure*}[t]
    \centering
  \includegraphics[width=1\textwidth]{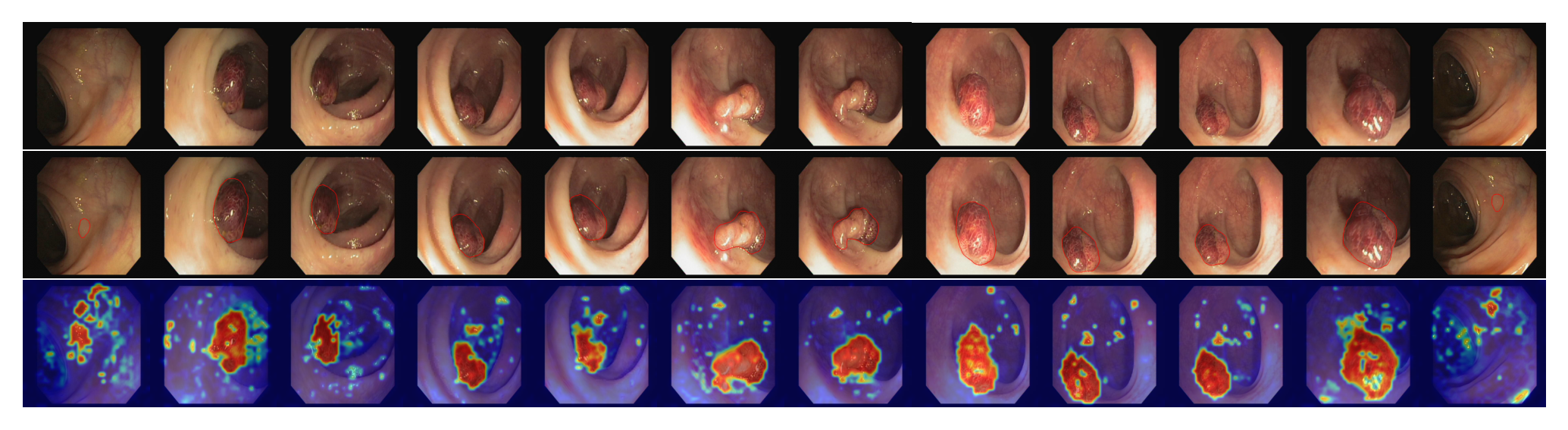}
    \caption{Anomaly score maps for the data subset colon. The first row represents the input, and we circle the anomaly regions in the second row. The last row presents the segmentation results from AnomalyCLIP.}
    \label{}
\end{figure*}

\begin{figure*}[t]
    \centering
  \includegraphics[width=1\textwidth]{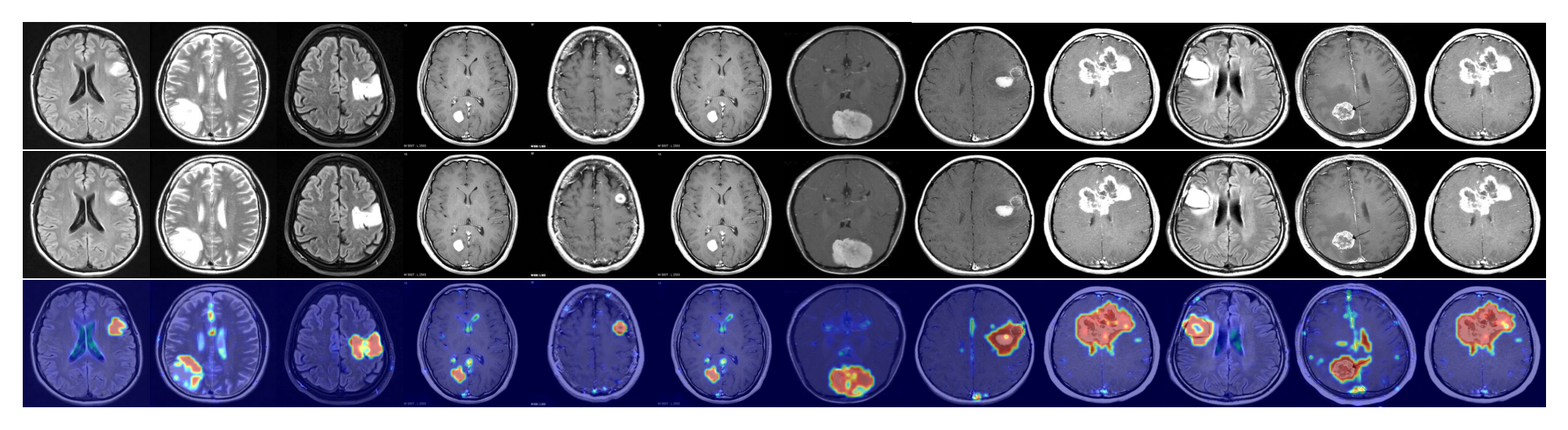}
    \caption{Anomaly score maps for the data subset brain. The first row represents the input, and we circle the anomaly regions in the second row. The last row presents the segmentation results from AnomalyCLIP.}

    \label{}
\end{figure*}

\section{Fine-grained ZSAD performance}
\label{Class-level performance}
In this section, we present the fine-grained data subset-level ZSAD performance in details.

\begin{table}[]
\centering
\caption{Fine-grained data-subset-wise performance comparison (AUROC) for anomaly segmentation on  MVTec AD.}
\tiny
\begin{tabular}{cccccccc}
\toprule
\textbf{Object name}  & CLIP & CLIP-AC & WinCLIP & VAND & CoOp & AnomalyCLIP  \\
\hline
Carpet      & 11.5  & 10.7   & 95.4  & 98.4  &  6.7   & 98.8 \\
Bottle      & 17.5  & 23.3   & 89.5   & 83.4  & 23.1   & 90.4\\
Hazelnut    & 25.2  & 34.0   & 94.3   & 96.1  & 30.2   & 97.1\\
Leather     &  9.9  &  5.6   & 96.7  & 99.1  & 11.7   & 98.6\\
Cable       & 37.4  & 37.5   & 77.0   & 72.3  & 49.7   & 78.9\\
Capsule     & 50.9  & 49.1   & 86.9   & 92.0  & 35.5   & 95.8\\
Grid        &  8.7  & 11.9   & 82.2   & 95.8  &  7.8   & 97.3\\
Pill        & 55.8  & 60.8   & 80.0   & 76.2  & 46.5   & 92  \\
Transistor  & 51.1  & 48.5   & 74.7   & 62.4  & 50.1   & 71  \\
Metal\_nut  & 43.9  & 53.6   & 61.0   & 65.4  & 49.3   & 74.4\\
Screw       & 80.1  & 76.4   & 89.6   & 97.8  & 17.0   & 97.5\\
Toothbrush  & 36.3  & 35.0   & 86.9   & 95.8  & 64.9   & 91.9\\
Zipper      & 51.5  & 44.7   & 91.6   & 91.1  & 33.4   & 91.4\\
Tile        & 49.9  & 39.1   & 77.6   & 92.7  & 41.7   & 94.6\\
Wood        & 45.7  & 42.4   & 93.4   & 95.8  & 31.4   & 96.5\\ \hline
Mean        & 38.4  & 38.2   & 85.1   & 87.6  & 33.3   & 91.1 \\
\bottomrule
\end{tabular}%

\end{table}

\begin{table}[]
\centering
\caption{Fine-grained data-subset-wise performance comparison (PRO) for anomaly segmentation on MVTec AD.}
\tiny
\begin{tabular}{cccccccc}
\toprule
\textbf{Object name}  & CLIP & CLIP-AC & WinCLIP & VAND & CoOp & AnomalyCLIP  \\
\hline
Carpet      &  2.9  &  1.9   & 84.1  & 48.5  &  0.5  & 90.1 \\
Bottle      &  1.4  &  4.9   & 76.4  & 45.6  &  4.5  & 80.9\\
Hazelnut    &  2.8  &  9.4   & 81.6  & 70.3  &  4.7  & 92.4\\
Leather     &  0.2  &  0.0   & 91.1  & 72.4  &  1.8  & 92.2 \\
Cable       &  7.3  &  6.9   & 42.9  & 25.7  & 12.2  & 64.4\\
Capsule     & 13.2  & 14.9   & 62.1  & 51.3  &  5.7  & 87.2\\
Grid        &  0.9  &  2.4   & 57.0  & 31.6  &  1.0  & 75.6\\
Pill        &  6.0  &  8.2   & 65.0  & 65.4  &  3.2  & 88.2\\
Transistor  & 15.3  & 11.2   & 43.4  & 21.3  &  9.3  & 58.1\\
Metal\_nut  &  2.9  & 10.3   & 31.8  & 38.4  &  7.0  & 71.0  \\
Screw       & 57.8  & 56.2   & 68.5  & 67.1  &  6.4  & 88.0  \\
Toothbrush  &  5.8  &  5.2   & 67.7  & 54.5  & 16.6  & 88.5\\
Zipper      & 17.7  & 15.2   & 71.7  & 10.7  & 11.6  & 65.3\\
Tile        & 21.5  & 16.3   & 51.2  & 26.7  & 10.1  & 87.6 \\
Wood        & 13.7  & 10.3   & 74.1  & 31.1  &  5.1  & 91.2\\ \hline
Mean        & 11.3  & 11.6   & 64.6  & 44.0  &  6.7  & 81.4 \\
\bottomrule
\end{tabular}%
\end{table}

\begin{table}[]
\centering
\caption{Fine-grained data-subset-wise performance comparison (AUROC) for anomaly classification on MVTec AD.}
\tiny
\begin{tabular}{cccccccc}
\toprule
\textbf{Object name}  & CLIP & CLIP-AC & WinCLIP & VAND & CoOp & AnomalyCLIP  \\
\hline
Carpet      & 96    & 93.1   &100.0   & 99.5  & 99.9   & 100.0  \\
Bottle      & 45.9  & 46.1   &99.2    & 92.0  & 87.7   &  89.3 \\
Hazelnut    & 88.7  & 91.1   &93.9    & 89.6  & 93.5   &  97.2 \\
Leather     & 99.4  & 99.5   &100.0   & 99.7  & 99.9   &  99.8 \\
Cable       & 58.1  & 46.6   &86.5    & 88.4  & 56.7   &  69.8 \\
Capsule     & 71.4  & 68.8   &72.9    & 79.9  & 81.1   &  89.9 \\
Grid        & 72.5  & 63.7   &98.8    & 86.3  & 94.7   &  97.0   \\
Pill        & 73.6  & 73.8   &79.1    & 80.5  & 78.6   &  81.8  \\
Transistor  & 48.8  & 51.2   &88.0    & 80.8  & 92.2   &  92.8 \\
Metal\_nut  & 62.8  & 63.4   &97.1    & 68.4  & 85.3   &  93.6 \\
Screw       & 78.2  & 66.7   &83.3    & 84.9  & 88.9   &  81.1 \\
Toothbrush  & 73.3  & 89.2   &88.0    & 53.8  & 77.5   &  84.7 \\
Zipper      & 60.1  & 36.1   &91.5    & 89.6  & 98.8   &  98.5 \\
Tile        & 88.5  & 89.0   &100.0   & 99.9  & 99.7   & 100.0  \\
Wood        & 94    & 94.9   &99.4    & 99.0  & 97.7   &  96.8 \\ \hline
Mean        & 74.1  & 71.5   &91.8    & 86.1  & 88.8   &  91.5  \\
\bottomrule
\end{tabular}%
\end{table}

\begin{table}[]
\centering
\caption{Fine-grained data-subset-wise performance comparison (AP) on for anomaly classification MVTec AD.}
\tiny
\begin{tabular}{cccccccc}
\toprule
\textbf{Object name}  & CLIP & CLIP-AC & WinCLIP & VAND & CoOp & AnomalyCLIP  \\
\hline
Carpet      & 98.8  & 97.8   & 100.0 & 99.8  & 100.0   &  100.0    \\
Bottle      & 78.9  & 79.8   & 99.8  & 97.7  &  96.4   &   97.0  \\
Hazelnut    & 94.6  & 95.9   & 96.9  & 94.8  &  96.7   &   98.6\\
Leather     & 99.8  & 99.8   & 100.0 & 99.9  & 100.0   &   99.9\\
Cable       & 70.8  & 64.3   & 91.2  & 93.1  &  69.4   &   81.4\\
Capsule     & 92.1  & 90.9   & 91.5  & 95.5  &  95.7   &   97.9\\
Grid        & 87.1  & 83.9   & 99.6  & 94.9  &  98.1   &   99.1 \\
Pill        & 93.4  & 93.6   & 95.7  & 96.0  &  94.2   &   95.4\\
Transistor  & 48.1  & 49.9   & 87.1  & 77.5  &  90.2   &   90.6\\
Metal\_nut  & 87.7  & 89.2   & 99.3  & 91.9  &  96.3   &   98.5\\
Screw       & 91.4  & 86.6   & 93.1  & 93.6  &  96.2   &   92.5\\
Toothbrush  & 90.7  & 96.0   & 95.6  & 71.5  &  90.4   &   93.7\\
Zipper      & 87.4  & 73.9   & 97.5  & 97.1  &  99.7   &   99.6\\
Tile        & 95.9  & 96.2   & 100.0 & 100.0 &  99.9   &  100.0   \\
Wood        & 97.9  & 98.3   & 99.8  & 99.7  &  99.4   &   99.2\\ \hline
Mean        & 87.6  & 86.4   & 96.5  & 93.5  &  94.8   &   96.2 \\
\bottomrule
\end{tabular}%
\end{table}

\begin{table}[]
\centering
\caption{Fine-grained data-subset-wise performance comparison (AUROC) for anomaly segmentation on VisA.}
\tiny
\begin{tabular}{ccccccc}
\toprule
\textbf{Object name} & CLIP & CLIP-AC & WinCLIP & VAND & CoOp & AnomalyCLIP  \\ \hline
Candle      & 33.6  & 50.0   & 88.9  & 97.8  & 16.3  & 98.8 \\
Capsules    & 56.8  & 61.5   & 81.6  & 97.5  & 47.5  & 95.0   \\
Cashew      & 64.5  & 62.5   & 84.7  & 86.0  & 32.5  & 93.8  \\  
Chewinggum  & 43.0  & 56.5   & 93.3  & 99.5  &  3.4  & 99.3  \\
Fryum       & 45.6  & 62.7   & 88.5  & 92.0  & 21.7  & 94.6  \\
Macaroni1   & 20.3  & 22.9   & 70.9  & 98.8  & 36.8  & 98.3  \\
Macaroni2   & 37.7  & 28.8   & 59.3  & 97.8  & 27.5  & 97.6  \\
Pcb1        & 57.8  & 51.6   & 61.2  & 92.7  & 19.8  & 94.1  \\
Pcb2        & 34.7  & 38.4   & 71.6  & 89.7  & 22.9  & 92.4  \\
Pcb3        & 54.6  & 44.6   & 85.3  & 88.4  & 18.0  & 88.4  \\
Pcb4        & 52.1  & 49.9   & 94.4  & 94.6  & 14.0  & 95.7  \\
Pipe\_fryum & 58.7  & 44.7   & 75.4  & 96.0  & 29.2  & 98.2  \\ \hline
Mean        & 46.6  & 47.8   & 79.6  & 94.2  & 24.2  & 95.5  \\
\bottomrule
\end{tabular}%
\end{table}

\begin{table}[]
\centering
\caption{Fine-grained data-subset-wise performance comparison (PRO) for anomaly segmentation on VisA.}
\tiny
\begin{tabular}{ccccccc}
\toprule
\textbf{Object name} & CLIP & CLIP-AC & WinCLIP & VAND & CoOp & AnomalyCLIP  \\ \hline
Candle      &  3.6  &  6.0   & 83.5  & 92.5  &  1.1  & 96.2\\
Capsules    & 15.8  & 22.4   & 35.3  & 86.7  & 18.4  & 78.5 \\
Cashew      &  9.6  & 10.9   & 76.4  & 91.7  &  1.7  & 91.6 \\
Chewinggum  & 17.8  & 30.2   & 70.4  & 87.3  &  0.1  & 91.2 \\
Fryum       & 12.1  & 29.3   & 77.4  & 89.7  &  2.6  & 86.8 \\
Macaroni1   &  8.1  & 13.4   & 34.3  & 93.2  & 18.1  & 89.8 \\
Macaroni2   & 20.9  & 18.4   & 21.4  & 82.3  &  2.7  & 84.2 \\
Pcb1        & 11.7  & 12.5   & 26.3  & 87.5  &  0.1  & 81.7 \\
Pcb2        & 12.8  & 13.9   & 37.2  & 75.6  &  0.7  & 78.9 \\
Pcb3        & 31.7  & 23.6   & 56.1  & 77.8  &  0.0  & 77.1 \\
Pcb4        & 17.1  & 20.3   & 80.4  & 86.8  &  0.0  & 91.3 \\
Pipe\_fryum & 16.7  &  6.0   & 82.3  & 90.9  &  0.6  & 96.8 \\ \hline
Mean        & 14.8  & 17.3   & 56.8  & 86.8  &  3.8  & 87.0   \\
\bottomrule
\end{tabular}%
\end{table}

\begin{table}[]
\centering
\caption{Fine-grained data-subset-wise performance comparison (AUROC) for anomaly classification on VisA.}
\tiny
\begin{tabular}{ccccccc}
\toprule
\textbf{Object name} & CLIP & CLIP-AC & WinCLIP & VAND & CoOp & AnomalyCLIP  \\ \hline
Candle      & 37.9  & 33.0   & 95.4   & 83.8  & 46.2  & 79.3 \\
Capsules    & 69.7  & 75.3   & 85.0  & 61.2  & 77.2  & 81.5 \\
Cashew      & 69.1  & 72.7   & 92.1  & 87.3  & 75.7  & 76.3 \\
Chewinggum  & 77.5  & 76.9   & 96.5  & 96.4  & 84.9  & 97.4 \\
Fryum       & 67.2  & 60.9   & 80.3  & 94.3  & 80.0  & 93.0 \\
Macaroni1   & 64.4  & 67.4   & 76.2  & 71.6  & 53.6  & 87.2 \\
Macaroni2   & 65    & 65.7   & 63.7  & 64.6  & 66.5  & 73.4 \\
Pcb1        & 54.9  & 43.9   & 73.6  & 53.4  & 24.7  & 85.4 \\
Pcb2        & 62.6  & 59.5   & 51.2  & 71.8  & 44.6  & 62.2 \\
Pcb3        & 52.2  & 49.0   & 73.4  & 66.8  & 54.4  & 62.7 \\
Pcb4        & 87.7  & 89.0   & 79.6  & 95.0  & 66.0  & 93.9 \\
Pipe\_fryum & 88.8  & 86.4   & 69.7  & 89.9  & 80.1  & 92.4 \\ \hline
Mean        & 66.4  & 65.0   & 78.1  & 78.0  & 62.8  & 82.1 \\
\bottomrule
\end{tabular}%
\end{table}

\begin{table}[]
\centering
\caption{Fine-grained data-subset-wise performance comparison (AP) for anomaly classification on VisA.}
\tiny
\begin{tabular}{ccccccc}
\toprule
\textbf{Object name} & CLIP & CLIP-AC & WinCLIP & VAND & CoOp & AnomalyCLIP  \\ \hline
Candle      & 42.9  & 40.0   & 95.8  & 86.9  & 52.9  & 81.1    \\
Capsules    & 81.0  & 84.3   & 90.9  & 74.3  & 85.3  & 88.7 \\
Cashew      & 83.4  & 86.1   & 96.4  & 94.1  & 87.1  & 89.4 \\
Chewinggum  & 90.4  & 90.2   & 98.6  & 98.4  & 93.1  & 98.9 \\
Fryum       & 82.0  & 76.6   & 90.1  & 97.2  & 90.2  & 96.8   \\
Macaroni1   & 56.8  & 58.7   & 75.8  & 70.9  & 52.3  & 86.0   \\
Macaroni2   & 65.0  & 65.8   & 60.3  & 63.2  & 62.2  & 72.1 \\
Pcb1        & 56.9  & 48.4   & 78.4  & 57.2  & 36.0  & 87.0   \\
Pcb2        & 63.2  & 59.8   & 49.2  & 73.8  & 47.3  & 64.3 \\
Pcb3        & 53.0  & 47.6   & 76.5  & 70.7  & 54.8  & 70.0   \\
Pcb4        & 88.0  & 90.6   & 77.7  & 95.1  & 66.3  & 94.4 \\
Pipe\_fryum & 94.6  & 93.7   & 82.3  & 94.8  & 89.7  & 96.3 \\ \hline
Mean        & 71.5  & 70.1   & 81.2  & 81.4  & 68.1  & 85.4 \\
\bottomrule
\end{tabular}%
\end{table}

\begin{table}[]
\centering
\caption{Fine-grained data-subset-wise performance comparison (AUROC) for anomaly segmentation on MPDD.}
\tiny
\begin{tabular}{ccccccc}
\toprule
\textbf{Object name} & CLIP & CLIP-AC & WinCLIP & VAND & CoOp & AnomalyCLIP  \\ \hline
Bracket\_black & 85.3  & 86.4  & 57.8  & 96.3  &  9.3   & 95.7  \\
Bracket\_brown & 26.9  & 31.5  & 72.2  & 86.2  & 20.2   & 94.4   \\
Bracket\_white & 83.5  & 77.4  & 79.5  & 99.0  &  8.3   & 99.8 \\
Connector      & 56.5  & 52.9  & 79.0  & 90.6  &  7.6   & 97.2 \\
Metal\_plate   & 64.3  & 52.5  & 92.6  & 93.1  & 14.1   & 93.8 \\
Tubes          & 56.4  & 51.5  & 77.6  & 99.1  & 33.2   & 98.1 \\ \hline
Mean           & 62.1  & 58.7  & 76.4  & 94.1  & 15.4   & 96.5 \\
\bottomrule
\end{tabular}%
\end{table}

\begin{table}[]
\centering
\caption{Fine-grained data-subset-wise performance comparison (PRO) for anomaly segmentation on MPDD.}
\tiny
\begin{tabular}{ccccccc}
\toprule
\textbf{Object name} & CLIP & CLIP-AC & WinCLIP & VAND & CoOp & AnomalyCLIP  \\ \hline
Bracket\_black & 62.6  & 58.9  & 43    & 89.7  &  1.5   & 85.2  \\
Bracket\_brown &  2.8  &  4.0  & 25.0  & 70.3  &  0.4   & 77.7 \\
Bracket\_white & 47.9  & 41.6  & 57.6  & 93.1  &  0.0   & 98.8 \\
Connector      & 22.8  & 20.2  & 44.6  & 74.5  &  0.0   & 89.8 \\
Metal\_plate   & 31.5  & 27.0  & 78.2  & 74.5  &  0.2   & 86.9 \\
Tubes          & 30.4  & 22.9  & 44.7  & 96.9  & 11.5   & 93.6 \\ \hline
Mean           & 33.0  & 29.1  & 48.9  & 83.2  &  2.3   & 88.7 \\
\bottomrule
\end{tabular}%
\end{table}

\begin{table}[]
\centering
\caption{Fine-grained data-subset-wise performance comparison (AUROC) for anomaly classification on MPDD.}
\tiny
\begin{tabular}{ccccccc}
\toprule
\textbf{Object name} & CLIP & CLIP-AC & WinCLIP & VAND & CoOp & AnomalyCLIP  \\ \hline
Bracket\_black & 32.4  & 32.8  & 41.5  & 66.1  &  36.9   & 67.3    \\
Bracket\_brown & 50.9  & 57.9  & 48.6  & 64.0  &  43.9   & 62.2 \\
Bracket\_white & 45.4  & 42.6  & 40.2  & 79.6  &  48.9   & 64.9   \\
Connector      & 75    & 76.2  & 79.3  & 78.8  &  38.3   & 86.9 \\
Metal\_plate   & 34.9  & 54.8  & 93.4  & 53.8  &  77.0   & 85.2 \\
Tubes          & 87.3  & 72.8  & 78.7  & 95.9  &  85.4   & 95.5   \\ \hline
Mean           & 54.3  & 56.2  & 63.6  & 73.0  &  55.1   & 77.0 \\
\bottomrule
\end{tabular}%
\end{table}

\begin{table}[]
\centering
\caption{Fine-grained data-subset-wise performance comparison (AP) for anomaly classification on MPDD.}
\tiny
\begin{tabular}{ccccccc}
\toprule
\textbf{Object name} & CLIP & CLIP-AC & WinCLIP & VAND & CoOp & AnomalyCLIP  \\ \hline
Bracket\_black & 47.8  & 48.6  & 56.9  & 71.7  &  50.0   & 72.9  \\
Bracket\_brown & 66.2  & 72.0  & 69.5  & 79.0  &  65.7   & 80.8   \\
Bracket\_white & 51.2  & 47.3  & 45.1  & 82.3  &  57.5   & 68.5 \\
Connector      & 62.2  & 61.4  & 61.3  & 71.8  &  26.4   & 76.8 \\
Metal\_plate   & 70.6  & 78.5  & 97.6  & 78.3  &  92.0   & 94.7 \\
Tubes          & 94.4  & 88.2  & 89.1  & 98.1  &  93.6   & 98.1 \\ \hline
Mean           & 65.4  & 66.0  & 69.9  & 80.2  &  64.2   & 82.0 \\
\bottomrule
\end{tabular}%
\end{table}

\end{document}

%% file: math_commands.tex
\usepackage{amsmath,amsfonts,bm}

\def\eqref#1{equation~\ref{#1}}
\def\1{\bm{1}}

\DeclareMathAlphabet{\mathsfit}{\encodingdefault}{\sfdefault}{m}{sl}
\SetMathAlphabet{\mathsfit}{bold}{\encodingdefault}{\sfdefault}{bx}{n}

\def\sG{{\mathbb{G}}}

\newcommand{\R}{\mathbb{R}}